\pdfoutput=1

\documentclass[11pt]{article}

\usepackage[]{acl}

\usepackage{times}
\usepackage{latexsym}

\usepackage[T1]{fontenc}

\usepackage[utf8]{inputenc}

\usepackage{microtype}
\usepackage{makecell}
\usepackage{inconsolata}
\usepackage{times}
\usepackage{pifont}
\usepackage{url}
\usepackage{latexsym}
\usepackage{graphicx}
\usepackage{subfigure}
\usepackage{booktabs}
\usepackage{amsmath}
\usepackage{amsthm}
\usepackage{enumitem}
\usepackage{color,amsfonts}
\usepackage{multirow,array}
\usepackage{tikz}
\usepackage{pgfplots}
\usepackage{algpseudocode}
\usepackage{amssymb,mathtools}
\usepackage[linesnumbered,ruled,vlined]{algorithm2e}
\usepackage{cleveref}
\usepackage{amssymb}
\usepackage{nomencl}
\usepackage{etoolbox}
\usepackage{tcolorbox}
\usepackage{caption}
\theoremstyle{definition}
\usepackage{array}
\usepackage{multirow}
\usepackage{booktabs}
\usepackage{ragged2e}
\usepackage{makecell}
\usepackage{graphicx}
\usepackage{tabularx}
\usepackage{siunitx}
\usepackage[table]{xcolor}
\usepackage{textcomp} 
\newcommand{\ph}[1]{\texttt{\textlangle #1\textrangle}}
\definecolor{lightblue}{rgb}{0.93,0.95,1.0}
\definecolor{lightgreen}{rgb}{0.9,1.0,0.9}
\definecolor{lightyellow}{rgb}{1.0,1.0,0.9}
\newcommand{\cmark}{\textcolor{green!80!black}{\ding{51}}}
\newcommand{\xmark}{\textcolor{red!80}{\ding{55}}}
\newcolumntype{P}[1]{>{\centering\arraybackslash}p{#1}}
\newcolumntype{M}[1]{>{\centering\arraybackslash}m{#1}}

\makeatletter
\patchcmd{\@algocf@start}
  {-1.5em}
  {0pt}
  {}{}
\makeatother

\usepackage{lipsum}
\usepackage{enumitem}
\usepackage{listings}
\usepackage{minted}  
\usepackage[dvipsnames]{xcolor}

\usepackage{amsthm,amsmath,amsfonts,bm,xspace}
\usepackage{color}

\theoremstyle{definition}
\usepackage{xspace}
\newcommand{\benchname}{Clinic\-Num\-Rob\-Bench\xspace}

\title{How Robust Are Large Language Models for Clinical Numeracy? \\ An Empirical Study on Numerical Reasoning Abilities in Clinical Contexts} 

\author{Minh-Vuong Nguyen$^\diamondsuit$ \enskip Fatemeh Shiri$^\diamondsuit$ \enskip Zhuang Li$^\diamondsuit$ \enskip Karin Verspoor$^\diamondsuit$ \\
  $^\diamondsuit$ School of Computing Technologies, RMIT University, Australia \\
   \texttt{\{vuong.nguyen2, first.last\}@rmit.edu.au} \\
}

\begin{document}
\maketitle

\begin{abstract}

Large Language Models (LLMs) are increasingly being explored for clinical question answering and decision support, yet safe deployment critically requires reliable handling of patient measurements in heterogeneous clinical notes. Existing evaluations of LLMs for \emph{clinical numerical reasoning} provide limited operation-level coverage, restricted primarily to arithmetic computation, and rarely assess the robustness of numerical understanding across clinical note formats. We introduce \textbf{\benchname},
a benchmark of 1{,}624 context--question instances with ground-truth answers that evaluates four main types of clinical numeracy: \emph{value retrieval}, \emph{arithmetic computation}, \emph{relational comparison}, and \emph{aggregation}. To stress-test robustness, \benchname presents longitudinal MIMIC-IV vital-sign records in three semantically equivalent representations, including a real-world note-style variant derived from the Open Patients dataset, and instantiates queries using 42 question templates. Experiments on 17 LLMs show that \emph{value retrieval} is generally strong, with most models exceeding 85\% accuracy, while \emph{relational comparison} and \emph{aggregation} remain challenging, with some models scoring below 15\%. Fine-tuning on medical data can reduce numeracy relative to base models by over 30\%, and performance drops under note-style variation indicate LLM sensitivity to format. \benchname offers a rigorous testbed for clinically reliable numerical reasoning\footnote{Code and data URL are available on
\url{https://github.com/MinhVuong2000/ClinicNumRobBench}}.

\end{abstract}

\section{Introduction}

Large Language Models (LLMs) have exhibited 
rapid exploration of medical applications such as clinical question answering and decision support \citep{bedi2025medhelm}. Yet safe clinical use critically requires LLMs to handle patient measurements reliably, which depends on both \emph{clinical numeracy}, executing basic numerical operations over clinical measurements, and \emph{clinical robustness}, preserving numerical correctness under semantically equivalent variations in how the same measurements are documented across clinical notes.

First, despite substantial progress on mathematical reasoning \citep{Shao2024DeepSeekMathPT}, prior numeracy evaluations show that LLMs still exhibit persistent weaknesses in fundamental operations, including \emph{value retrieval}, \emph{arithmetic computation}, \emph{relational comparison}, and \emph{aggregation} \citep{li2025exposing,mahendra2025evaluating}. Meanwhile, clinical text is numerically rich, with numbers appearing in most documents across widely used clinical datasets \citep{mahendra2024numbers}, and these same operations are repeatedly required in clinical workflows over patient measurements, making numerical reliability a prerequisite for safe deployment.

Second, clinical numeracy differs from canonical mathematical word problems with systematic symbolic structure. Clinical measurements are embedded in diverse and noisy contexts, and the same value may appear in structured fields, semi-structured templates, or free-text notes with varied surface forms (e.g., ``BP: 120/80'' vs.\ ``blood pressure of 120 over 80'') \citep{Rothman2008PerspectiveTR}. As a result, even when an LLM succeeds in one presentation, its numerical behavior may be brittle under semantically equivalent documentation variants.

A clinically meaningful evaluation of LLMs must therefore test both fine-grained numerical operation correctness and robustness under representational shifts where the underlying measurements remain unchanged. However, existing datasets do not jointly address these requirements. General numeracy benchmarks~\cite{mahendra2024numbers, mahendra2025evaluating,li2025exposing} are mostly non-clinical, while clinical benchmarks often cover only a subset of operations and omit robustness testing. 
For instance, MedCalc-Bench \citep{khandekar2024medcalc} is derived from patient notes but mainly evaluates \emph{arithmetic computation} and Electromyogram Table Mart (ETM) \citep{long2025emgllm} focuses on table-to-text diagnosis generation rather than operation-level clinical numeracy benchmarking. Neither consider robustness under note-style variation. A recent study \citep{gourabathina2025medium} examined LLMs' sensitivity to non-clinical input perturbation, with altered patient attributes causing inconsistent treatment recommendations and demographic disparities. While the work focuses on the fairness of LLMs with the change of non-clinical input, the robustness of LLMs on diverse clinical documentation formats remains unexplored, especially numerical reasoning being fundamental to clinical decision-making.

To address this gap, we introduce \textbf{\benchname}, a benchmark of 1{,}624 context--question instances for fine-grained clinical numeracy evaluation and robustness testing, together with a scalable synthetic construction pipeline. Grounded in real-world MIMIC-IV vital-sign records and demographics \citep{johnson2023mimic,johnson2023mimicorigin, johnson2024mimic}, 
our pipeline automatically constructs longitudinal patient contexts by sampling and assembling records, then renders the same underlying measurements in three semantically equivalent context representations, from structured formats to realistic natural-language variants. We instantiate 42 question templates to generate queries probing \emph{value retrieval}, \emph{arithmetic computation}, \emph{relational comparison}, and \emph{aggregation} operations, and compute ground-truth answers programmatically. Using \benchname, we evaluate 17 LLMs across three categories and report three findings in task accuracy, medical fine-tuning negative impact, and context robustness.


Our main contributions are:
\begin{itemize}\setlength\itemsep{0.2em}
    \item We introduce \textbf{\benchname}, a benchmark of 1{,}624 instances for evaluating clinical numeracy across 4 operations and robustness across 3 context representation formats.
    \item We propose a scalable construction pipeline grounded in MIMIC-IV longitudinal records that automatically builds the benchmark with minimal human effort.
    \item Experiments on 17 LLMs reveal persistent weaknesses in comparison and aggregation, potential numeracy degradation from medical fine-tuning, and strong sensitivity to clinical note representational shifts.
\end{itemize}

\section{Related Work}
\paragraph{Clinical numeracy evaluation.}
Most numeracy benchmarks evaluate LLMs using general mathematics questions, ranging from grade school arithmetic to Olympiad level problems \citep{cobbe2021training,hendrycksmath2021,li2025exposing}. Despite strong performance on difficult problems, LLMs can still fail simple operations such as comparison or multiplication \citep{li2025exposing,mahendra2025evaluating}. Clinical text is numerically rich, yet widely used medical benchmarks such as MedQA, MedXpertQA, and MedBullets \citep{jin2020disease,zuo2025medxpertqabenchmarkingexpertlevelmedical,chen-etal-2025-benchmarking} assess broad clinical knowledge and reasoning, making it hard to isolate numeracy. \citet{mahendra2024numbers} identified prevalent types of numerical information in clinical documents, highlighting the lack of systematic resources in the clinical domain for numerical understanding and reasoning. Recently targeted studies on medical calculations and table interpretation \citep{khandekar2024medcalc,long2025emgllm} focus on limited formulas or modalities and do not systematically test value retrieval, arithmetic computation, relational comparison, and aggregation.

\paragraph{Robustness to documentation variation.}
LLMs are sensitive to small changes in prompt wording or formatting \citep{zhao2021calibrateuseimprovingfewshot,Zhuo2024ProSAAA,Arora2025ExploringRO}. This has motivated robustness benchmarks such as GSM Plus \citep{Li2024GSMPlusAC}, FinBias \citep{Mehrotra2025UnmaskingBI}, and CARES \citep{Chen2025CARESCE}. In the medical domain, robustness studies mainly target knowledge-oriented adversarial prompts \citep{Ness2024MedFuzzET}, or non-clinical input perturbations affecting clinical recommendations \citep{gourabathina2025medium}, and rarely examine numerical robustness, even though clinical measurements appear in structured EHR fields, semi-structured templates, and free text notes. Reliable deployment therefore requires testing whether numerical accuracy holds under semantically equivalent documentation variants.

\section{\benchname}
\label{sec:benchmark}
\begin{figure*}[t]
    \centering
    \includegraphics[width=0.95\textwidth]{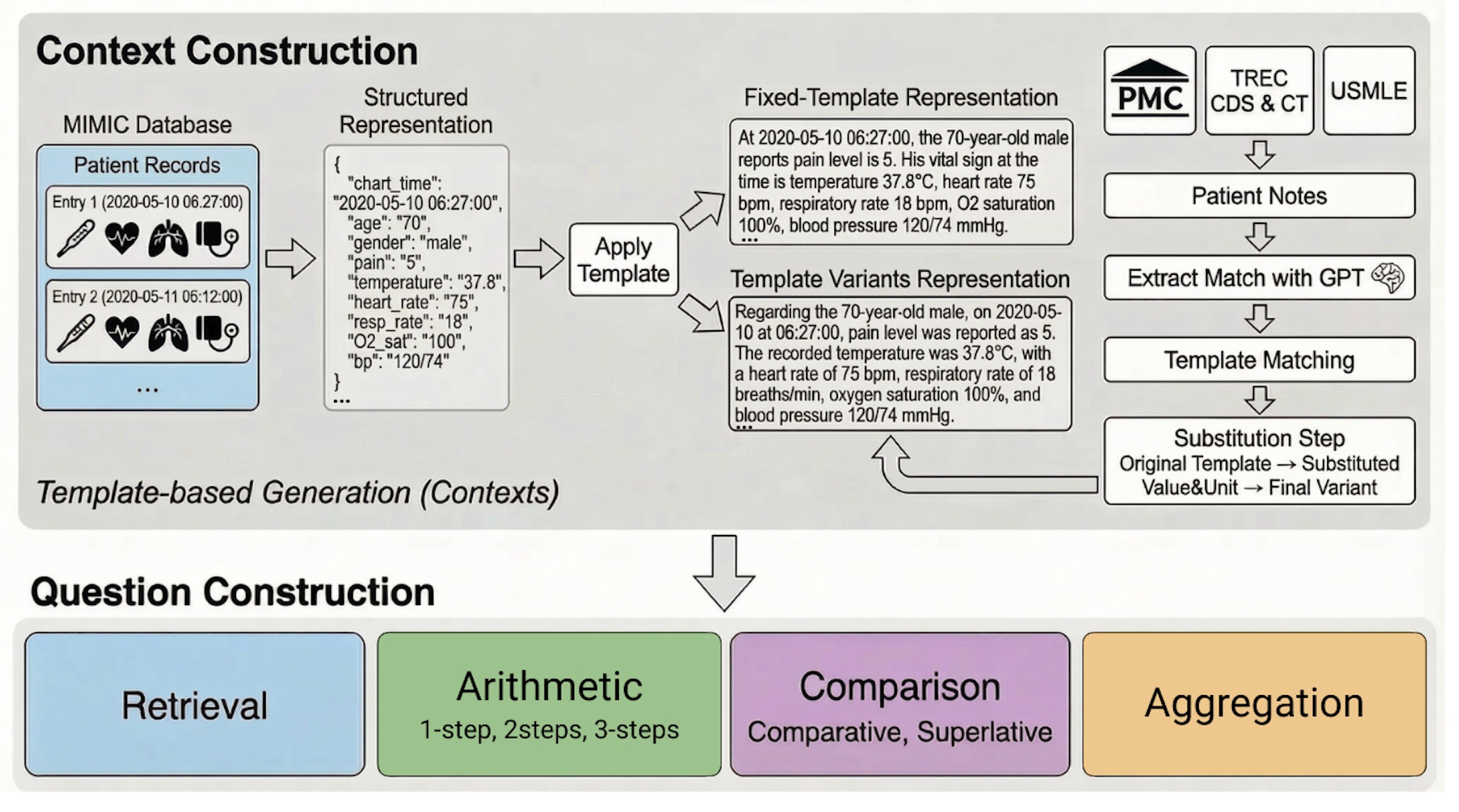}
    \caption{Overview of benchmark construction.}\label{fig:full_context_flow}
\end{figure*}

In this section, we present the \textbf{Clinic}al \textbf{Num}eracy \textbf{Rob}ustness \textbf{Bench}mark (\textbf{\benchname}) and its construction pipeline, which is used to evaluate numerical reasoning of LLMs in clinical settings and their robustness to semantically equivalent documentation variants. Each instance consists of a patient \emph{context}, a \emph{question}, and an exact ground-truth answer. \benchname integrates (i) multiple clinical context representations and (ii) template-based tasks that target core numerical operations. We then detail context construction, task and question generation, and dataset statistics. A detailed comparison with existing benchmarks, particularly the arithmetic-focused MedCalc-Bench, appears in Table~\ref{tab:benchmark_comparison_medcalc_detail} and Appendix~\ref{appdx:benchmarks_comparison}.

\subsection{Context Construction}
\label{subsec:context_preparation}

Real world clinical measurements appear in heterogeneous formats, from structured \emph{representations} (e.g., JSON, XML, tables) in EHR systems to unstructured natural language in clinical notes and physician narratives. Because documentation practices vary widely, LLM numeracy behavior may be sensitive to surface form changes. To measure robustness under such representational shifts, we construct three semantically equivalent context \emph{representations}: (i) a structured (JSON like) format, (ii) a fixed natural language template, and (iii) realistic note style variants derived from clinical notes.

We construct contexts from longitudinal vital-sign records and demographics, and render the same underlying measurements into the three semantically equivalent representations described above. Concretely, we (1) extract patient profiles from \texttt{hosp/patients} and \texttt{icu/icustays} in MIMIC-IV \citep{johnson2024mimic}, (2) link them to vital-sign records in MIMIC-IV ED \citep{johnson2023mimic}, 
(3) perform stratified sampling over age and gender to obtain 200 patients with 1--50 records each, and (4) instantiate each representation by filling structured and natural-language templates with the resulting values. Each record contains a timestamp, demographics, and clinical measurements, including vital signs and pain level.

\subsubsection{Structured Representation}
\label{context_template:json}

We serialize each record as a JSON-like key--value object and represent a patient context as a chronological list of these objects. Keys are fixed field names, including \texttt{chart\_time}, demographic fields, and vital-sign attributes shown in Figure~\ref{fig:full_context_flow}, and values are the corresponding measurements from the underlying records. This representation mirrors common structured EHR exports and tests whether LLMs can reason over structured clinical data without relying on natural-language cues.

\subsubsection{Fixed-template Representation}

Prior work has converted structured records into natural language form through verbalization \citep{nguyen2024carer, oliveira2025development}. We adopt the same idea to construct our second context representation. Using a single fixed template enables controlled comparison, since it holds the linguistic form largely constant while varying only the underlying values. This design helps isolate the effect of light linguistic wrapping, such as function words and sentence structure, and quantify robustness to modest representational shifts.

For each patient, we verbalize all records in chronological order. Each record is rendered using the fixed template below:

\begin{tcolorbox}[
  title=Record verbalized fixed template,
  fonttitle=\bfseries\small,
  boxrule=0.4pt, arc=1mm,
  left=1mm,right=1mm,top=0.6mm,bottom=0.6mm
]
\small
\raggedright
At \ph{charttime}, the \ph{age}-year-old \ph{gender} reports a pain level of \ph{pain}. Vital signs: temperature \ph{temperature}$^\circ$C, heart rate \ph{heartrate} bpm, respiratory rate \ph{resprate} breaths/min, O$_2$ saturation \ph{o2sat}\%, blood pressure \ph{sbp}/\ph{dbp} mmHg.
\end{tcolorbox}

For patients with multiple records, we concatenate the verbalized records using a newline character \texttt{\textbackslash n}.
For readability, the first record states age and gender explicitly, while subsequent records use pronouns such as ``his'' or ``her''. A complete example is shown in Figure~\ref{fig:full_context_flow}.

\subsubsection{Realistic Variant Templates Representation}
\label{subsubsec:variant_context}

Fixed templates cannot capture the linguistic diversity of real clinical documentation.
To approximate note-style variability while keeping the underlying measurements unchanged, we derive variant templates from real patient notes and then fill them with the same structured vital-sign values.

We construct these note-style variants as follows:
\begin{itemize}[leftmargin=*, itemsep=0.2em]
    \item \textbf{Collect candidate notes.} We source notes from the Open Patients dataset \citep{khandekar2024medcalc}, which contains 180K notes drawn from multiple clinical corpora, including \textbf{T}ext \textbf{RE}trieval \textbf{C}onference (TREC) Clinical Decision Support/Clinical Trials tracks, MedQA-USMLE, and PMC-Patients.
    \item \textbf{Filter for numerically rich notes.} We keep notes that contain at least five numbers and a \texttt{vital sign} keyword, yielding 15K candidates.
    \item \textbf{Extract vital-sign mentions.} Using \texttt{gpt-4.1-mini}, we extract five vital-sign attributes (temperature, heart rate, respiratory rate, oxygen saturation, and blood pressure). For each attribute, we output a JSON object containing the raw mention span, numeric value, and unit. Missing fields are recorded as \texttt{None}.
    \item \textbf{Validate extraction quality.} We require each extracted raw span to appear in the original note, and re-run extraction until this condition is satisfied. This yields 2{,}300 notes with validated mentions of all five vital signs.
    \item \textbf{Form variant templates.} For each validated note, we locate the extracted mention indices, take the minimal span covering all mentions, and replace each numeric value and unit with placeholders \texttt{\%v} and \texttt{\%m}.
    \item \textbf{Fill templates with record values.} We apply the resulting templates to structured vital-sign records by filling placeholders with record values, producing diverse note-style context variants.
    \item \textbf{Assemble patient contexts.} For each patient, we order the resulting note-style records chronologically and concatenate them to form the temporal narrative used as input for all benchmark tasks.
\end{itemize}

The extraction prompt and template categories with their frequencies are provided in the Appendix (Figures~\ref{fig:template_variant_categorization} and~\ref{fig:extraction_prompt}, respectively). An end-to-end pipeline
is shown in Figure~\ref{fig:full_context_flow}.

\subsection{Task Design and Question Generation}
\label{subsec:task_initialization}
\benchname evaluates four clinical numeracy operation types: \textbf{value retrieval}, \textbf{arithmetic computation}, \textbf{relational comparison}, and \textbf{aggregation}. We create instances for each operation using template based questions whose answers are computed deterministically from the underlying measurements. Each template specifies (i) the relevant record or records in a patient’s longitudinal context, (ii) the required numerical operation, and (iii) an exact ground truth answer. For retrieval, arithmetic computation, and aggregation tasks, answers are formatted as numeric values (integers or decimals rounded to two decimal places). For comparison tasks, answers are formatted as timestamp strings (\texttt{YYYY-MM-DD HH:MM}). 

To instantiate a template, we first select the relevant record(s) from the structured vital sign sequence according to the template’s constraints (for example by timestamp, by position, or by a threshold). We then fill the template slots (such as \verb|<charttime>|, \verb|<vital_parameter>|, or a threshold) with the corresponding values extracted from these records and compute the answer deterministically using the specified operation. Thresholds are selected based on their clinically meaningful cutoffs. The resulting filled question is paired with the same patient context rendered in each of the three context representations (\S\ref{subsec:context_preparation}). Table~\ref{tab:task-amount} reports the number of instances and templates per task type. Table~\ref{tab:question_templates} in the Appendix provides further details  by subtask. 
Table \ref{tab:task-type-alignment} depicts the alignment of benchmark tasks with clinical numerical information types \citep{mahendra2024numbers}.
\begin{table}[bt]
\centering
\resizebox{\columnwidth}{!}{
\begin{tabular}{l l r r}
\toprule
\textbf{Task}     & \textbf{Subtask} & \textbf{\#Templates} & \textbf{\#Questions} \\
\midrule
Retrieval         & --          & 6  & 240  \\
\midrule
\multirow{3}{*}{Arithmetic} & 1 step    & 4  & 200  \\
                             & 2 steps   & 4  & 200  \\
                             & 3 steps   & 2  & 200  \\
\midrule
\multirow{2}{*}{Comparison}  & Comparative & 11 & 191 \\
                             & Superlative & 12 & 192 \\
\midrule
Aggregation      & --          & 3  & 201  \\
\midrule
\textbf{Total}   &              & \textbf{42} & \textbf{1{,}624} \\
\bottomrule
\end{tabular}}
\caption{Number of templates and questions per task.}
\label{tab:task-amount}
\end{table}

\subsubsection{Value Retrieval}
\label{subsubsec:retrieval_tasks}

Value retrieval involves the location and extraction of a single measurement from the patient context given a timestamp. We generate queries over six vital parameters: temperature, heart rate, respiratory rate, oxygen saturation, systolic blood pressure, and diastolic blood pressure. To cover both early and later observations, we sample target records at fixed positions in the timeline (for example the first and fifth records when available) and fill the template with the corresponding measurement and timestamp. The ground truth answer is simply the value of the requested vital sign at that timestamp.

\subsubsection{Arithmetic Computation}
\label{subsubsec:calculation_tasks}

Arithmetic computation relates to clinically motivated arithmetic over one or more measurements in the context. We group questions into one-step, two-step and three-step subtasks, defined by the number of arithmetic operations involved. For each instance, we select the required record(s) by timestamp or position, substitute their values into the arithmetic expression defined by the template, and compute the exact numeric result as the answer (e.g., $\mathrm{MAP}=\mathrm{DBP}+\tfrac{1}{3}(\mathrm{SBP}-\mathrm{DBP})$, where MAP, SBP and DBP denote mean arterial pressure, systolic and diastolic blood pressure, respectively). The formulas utilized were derived from the intersection between vital sign information and the 50 most frequently accessed calculators available on MDCalc\footnote{\href{https://www.mdcalc.com/\#Popular}{MDCalc: https://www.mdcalc.com/\#Popular}}.

\subsubsection{Relational Comparison}
\label{subsubsec:comparison_tasks}

Relational comparison evaluates ordering and threshold reasoning over longitudinal measurements, including identifying extreme values. We design two query types: (i) \emph{comparative} queries that return the first timestamp at which a vital sign exceeds (\texttt{higher}) or falls below (\texttt{lower}) a threshold, and (ii) \emph{superlative} queries that return the timestamp of the maximum (\texttt{highest}) or minimum (\texttt{lowest}) value of a vital parameter. In each case, the template specifies the target vital sign and, for comparative queries, the threshold, and the ground truth answer is the timestamp (YYYY-MM-DD HH:MM) of the matching record.

\subsubsection{Aggregation}
\label{subsubsec:summary_tasks}

Aggregation requires combining information across the longitudinal context by counting how many records satisfy a clinical condition over time. Aggregation underlies clinical monitoring patterns such as counting occurrences of abnormal vitals. For each instance, we scan all records, count the number of matches (sometimes after computing a derived quantity like a shock index), and return this integer as the answer.




\begin{table*}[t]
\centering
\small
\setlength{\tabcolsep}{4pt}
\renewcommand{\arraystretch}{1.15}
\resizebox{0.95\textwidth}{!}{
\begin{tabular}{l c c |c c c c}
\toprule
\textbf{Benchmark} & \textbf{\#Inst.} & \textbf{Tasks} & \textbf{Ctx Rep.} &
\textbf{Ctx Tok.} & \textbf{Q Tok.} & \textbf{Root TTR} \\
\midrule
MedCalc-Bench & 1{,}047 & arithmetic & note-style & 613 & 41 & 10.71 \\
\midrule
\benchname{} & 1{,}624 & retrieval; & structured & 745 & 23 & 4.93--5.49 \\
(Ours) &  & arithmetic; & fixed template & 551 & 23 & 4.28--5.51 \\
 &  & comparison; aggregation & note-style variant & 1480 & 23 & 12.56--12.87 \\
\bottomrule
\end{tabular}
}
\caption{Comparison between \benchname{} and MedCalc-Bench. Each \benchname{} row corresponds to one context representation (structured, fixed template, or note-style variant). \textbf{Ctx Tok.} and \textbf{Q Tok.} denote the average number of context and question tokens, respectively. \textbf{Root TTR} denotes the lexical diversity range of the context across all four tasks (root type–token ratio); see Appendix~\ref{appdx:lexical_diversity} for details.}
\label{tab:benchmark_comparison_medcalc_detail}
\end{table*}


\subsection{Data Comparison and Analysis}
Table~\ref{tab:task-amount} summarizes \benchname{}, which comprises 42 templates and 1,624 instances spanning four clinical numeracy operations: retrieval (240), arithmetic (600), comparison (383), and aggregation (201). Arithmetic is evenly distributed across 1-, 2-, and 3-step problems (200 instances each), and comparison is balanced between comparative and superlative queries (191 vs.\ 192), enabling fine-grained analysis of threshold versus extremum reasoning.

Table~\ref{tab:benchmark_comparison_medcalc_detail} contrasts \benchname{} with MedCalc-Bench, the closest benchmark to ours for clinical numeracy. MedCalc-Bench focuses on arithmetic in a single note-style format, whereas \benchname{} expands coverage to retrieval, comparison, and aggregation, and provides three semantically equivalent context representations for robustness analysis. The structured and fixed-template representations yield shorter contexts (551--745 context tokens) and lower lexical diversity (Root TTR 4.28--5.51), serving as controlled baselines. In contrast, the note-style variant is substantially longer (1,480 context tokens) and more lexically diverse (Root TTR 12.56--12.87). Compared with MedCalc-Bench (context tokens 613; Root TTR 10.71), our note-style variant also exhibits greater lexical diversity. Since measurements and questions are held fixed across representations, performance differences can be more directly attributed to representational shift.



\section{Experiments}
\newcommand{\accdecrease}[1]{\hspace{0.15em}{\fontsize{8}{1}\selectfont \textcolor{red}{#1}}}

\newcommand{\accincrease}[1]{\hspace{0.15em}{\fontsize{8}{1}\selectfont \color{Green} #1}}

\providecommand{\rmark}{\textsuperscript{\ttfamily\scriptsize R}}

\newcolumntype{N}{S[table-format=3.2]}
\newcolumntype{D}{S[table-format=3.2, table-space-text-post=\accdecrease{-24.00}]}

\newcommand{\mc}[1]{\multicolumn{1}{c}{#1}}

\begin{table*}[t!]
\centering
\resizebox{\textwidth}{!}{
\begin{tabular}{@{}l c rrr rrr rrr rrr@{}}
\toprule
\multirow{2}{*}{\textbf{LLMs}} & \multirow{2}{*}{\textbf{Size}} &
\multicolumn{3}{c}{\textbf{Retrieval}} &
\multicolumn{3}{c}{\textbf{Arithmetic}} &
\multicolumn{3}{c}{\textbf{Comparison}} &
\multicolumn{3}{c}{\textbf{Aggregation}} \\
\cmidrule(lr){3-5}\cmidrule(lr){6-8}\cmidrule(lr){9-11}\cmidrule(lr){12-14}
& & Struct. & Fixed & Variant & Struct. & Fixed & Variant & Struct. & Fixed & Variant & Struct. & Fixed & Variant \\
\midrule

\multicolumn{14}{@{}l}{\textbf{Medical}} \\
\addlinespace[0.2em]
\texttt{medgemma-it} & 4B &
97.50 & 93.75\accdecrease{-3.75} & 90.83\accdecrease{-6.17} &
71.33 & 73.17\accincrease{+1.84} & 68.83\accdecrease{-2.50} &
1.56  & 2.35\accincrease{+0.79}  & 1.83\accincrease{+0.27} &
15.92 & 21.87\accincrease{+5.95} & 15.42\accdecrease{-0.50} \\
\texttt{MediPhi} & 4B &
94.58 & 93.33\accdecrease{-1.25} & 86.67\accdecrease{-7.91} &
22.00 & 18.67\accdecrease{-3.33} & 16.50\accdecrease{-5.50} &
29.50 & 29.76\accdecrease{-0.26} & 20.89\accdecrease{-8.61} &
15.92 & 26.37\accincrease{+10.45} & 13.93\accdecrease{-1.99} \\
\texttt{Meditron3-Qwen2.5} & 7B &
98.33 & 98.33\accincrease{+0.00} & 90.00\accdecrease{-8.33} &
83.33 & 82.67\accdecrease{-0.66} & 73.50\accdecrease{-9.93} &
36.29 & 47.52\accincrease{+11.23} & 34.20\accdecrease{-2.09} &
50.25 & 51.74\accincrease{+1.49} & 44.28\accdecrease{-7.46} \\
\texttt{Meditron3-Llama} & 8B &
75.83 & 79.58\accincrease{+3.75} & 73.75\accdecrease{-2.08} &
43.33 & 53.50\accincrease{+10.17} & 46.50\accincrease{+3.15} &
2.35  & 4.18\accincrease{+1.83}  & 3.13\accincrease{+0.78} &
15.42 & 16.92\accincrease{+1.50} & 11.94\accdecrease{-3.48} \\
\texttt{UltraMedical} & 8B &
89.58 & 91.25\accincrease{+1.67} & 66.25\accdecrease{-23.33} &
66.67 & 71.33\accincrease{+4.66} & 42.67\accdecrease{-24.00} &
27.23 & 33.68\accincrease{+6.45} & 18.85\accdecrease{-8.38} &
31.34 & 30.85\accdecrease{-0.49} & 15.42\accdecrease{-15.92} \\
\texttt{Huatuo-o1} & 8B &
97.08 & 91.67\accdecrease{-5.41} & 82.92\accdecrease{-14.16} &
59.83 & 57.50\accdecrease{-2.33} & 52.17\accdecrease{-7.66} &
49.61 & 55.35\accincrease{+5.66} & 27.94\accdecrease{-21.67} &
35.32 & 35.32\accincrease{+0.00} & 25.87\accdecrease{-9.45} \\

\midrule
\multicolumn{14}{@{}l}{\textbf{Base}} \\
\addlinespace[0.2em]
\texttt{gemma-3-it} & 4B &
\textbf{100.00} & 97.08\accdecrease{-2.92} & 92.08\accdecrease{-9.92} &
79.17 & 81.33\accincrease{+2.16} & 73.00\accdecrease{-6.17} &
63.45 & 61.88\accdecrease{-1.75} & 40.47\accdecrease{-22.98} &
41.79 & 43.78\accincrease{+1.99} & 30.85\accdecrease{-10.94} \\
\texttt{Phi-3.5-mini-reasoning} & 4B &
98.33 & 98.75\accincrease{+0.42} & 89.17\accdecrease{-9.16} &
75.83 & 79.33\accincrease{+3.50} & 69.83\accdecrease{-6.00} &
53.26 & 57.70\accincrease{+4.44} & 46.82\accdecrease{-6.44} &
55.22 & 59.70\accincrease{+4.48} & 38.36\accdecrease{-16.86} \\
\texttt{Qwen-2.5} & 7B &
\textbf{100.00} & 99.17\accdecrease{-0.83} & 89.58\accdecrease{-10.42} &
89.17 & 90.33\accincrease{+1.16} & 82.33\accdecrease{-6.84} &
67.36 & 72.33\accincrease{+4.97} & 52.74\accdecrease{-14.62} &
60.20 & 58.21\accdecrease{-1.99} & 49.25\accdecrease{-10.95} \\
\texttt{llama-3.1} & 8B &
95.00 & 96.67\accincrease{+1.67} & 88.75\accdecrease{-6.25} &
60.83 & 70.50\accincrease{+9.67} & 66.33\accincrease{+5.50} &
18.36 & 32.12\accincrease{+13.67} & 25.51\accincrease{+7.15} &
15.92 & 29.85\accincrease{+13.93} & 20.90\accincrease{+4.98} \\

\midrule
\multicolumn{14}{@{}l}{\textbf{General}} \\
\addlinespace[0.2em]
\texttt{DeepSeek-R1-Distill} & 8B &
97.08 & 89.17\accdecrease{-7.91} & 85.00\accdecrease{-12.08} &
63.50 & 66.67\accincrease{+3.17} & 52.33\accdecrease{-11.17} &
63.97 & 61.10\accdecrease{-2.87} & 42.97\accdecrease{-21.00} &
42.79 & 34.33\accdecrease{-8.46} & 30.35\accdecrease{-12.44} \\
\texttt{Qwen3} & 8B &
99.58 & \textbf{100.00}\accincrease{+0.42} & 90.42\accdecrease{-9.16} &
94.67 & 91.50\accdecrease{-3.17} & 82.67\accdecrease{-12.00} &
71.28 & 74.93\accincrease{+3.65} & 59.61\accdecrease{-11.67} &
76.62 & 76.62\accincrease{+0.00} & 60.22\accdecrease{-16.40} \\
\texttt{GPT-4.1-mini} & N/A &
\textbf{100.00} & \textbf{100.00}\accincrease{+0.00} & 96.25\accdecrease{-3.75} &
96.67 & 96.33\accdecrease{-0.34} & 92.50\accdecrease{-4.17} &
\textbf{95.30} & 95.04\accdecrease{-0.26} & 83.03\accdecrease{-12.27} &
\textbf{82.59} & 82.09\accdecrease{-0.50} & 71.64\accdecrease{-10.95} \\
\texttt{Qwen3-reasoning} & 8B &
\textbf{100.00} & \textbf{100.00}\accincrease{+0.00} & 87.92\accdecrease{-12.08} &
97.17 & 97.50\accincrease{+0.33} & 94.33\accdecrease{-2.84} &
94.33 & \textbf{95.30}\accincrease{+0.97} & 77.28\accdecrease{-17.05} &
76.62 & \textbf{82.59}\accincrease{+5.97} & 63.18\accdecrease{-13.44} \\

\texttt{gemma-3-it} & 27B &
\textbf{100} & \textbf{100}\accincrease{+0.00} & 94.58\accdecrease{-5.42} &
95.20 & 96.75\accincrease{+1.55} & 92.83\accdecrease{-2.37} &
78.61 & 79.51\accincrease{+0.90} & 68.26\accdecrease{-10.35} &
60.15 & 58.12\accdecrease{-2.03} & 49.18\accdecrease{-10.97} \\

\texttt{llama-3.3} & 70B &
99.58 & \textbf{100}\accincrease{+0.42} & 96.25\accdecrease{-3.33} &
96.00 & 96.33\accincrease{+0.33} & 92.17\accdecrease{-3.83} &
93.74 & 95.04\accincrease{+1.30} & 79.37\accdecrease{-14.37} &
73.63 & 75.12\accincrease{+1.49} & 61.69\accdecrease{-11.94} \\

\texttt{GPT-5} & N/A &
\textbf{100} & \textbf{100}\accincrease{+0.00} & 98.75\accdecrease{-1.25} &
97.33 & \textbf{98.00}\accincrease{+0.67} & 93.5\accdecrease{-3.83} &
86.67 & 84.75\accdecrease{-1.92} & 72.22\accdecrease{-14.45} &
78.61 & 79.60\accincrease{+0.99} & 62.19\accdecrease{-16.42} \\
\bottomrule
\end{tabular}
}
\caption{Accuracy (\%) of LLMs on four clinical numeracy tasks. Struct., Fixed, and Variant denote the three context representations. \accincrease{+x.xx}, \accdecrease{-x.xx} indicates the accuracy increase and decrease relative to the Struct.\  score in the same row, respectively. The suffix ``-reasoning'' indicates that the model is evaluated with its reasoning mode enabled.}
\label{tab:main-result}
\vspace{-1em}
\end{table*}

We use \benchname to benchmark LLMs on four core numerical skills and their robustness across three clinical context representations. 

\paragraph{Evaluation.}

We report the accuracy for each task, defined as $\textrm{accuracy}=\frac{\textrm{match}}{\textrm{match} + \textrm{unmatch}}$.

For all the output types, the predicted answer should exactly match the ground truth answer. To ensure robust evaluation despite minor formatting variations in LLM numeric outputs, we apply the following post-processing steps: (1) we extract numeric values using the regular expression pattern \texttt{-?\textbackslash d*\textbackslash.?\textbackslash d+}, and (2) we normalize trailing zeros by removing $.0+$ suffixes. This preprocessing handles common LLM output variations, such as appending $.0$ to integers or including measurement units, while preserving the semantic correctness of predictions. Correctness requires the exact post-processed match with ground truth. 

\paragraph{Evaluated Models.}
We evaluated 17 widely used and efficient LLMs that span three categories: medical-domain models, their corresponding base models, and selected general-purpose LLMs.

\begin{itemize}[leftmargin=*,itemsep=0.2em]
\item \textbf{Medical LLMs:} MedGemma~\citep{sellergren2025medgemma}, MediPhi~\citep{corbeil2025modular}, Meditron3-Llama3.1~\citep{chen2024meditron}, Meditron3-Qwen2.5, UltraMedical~\citep{zhang2024ultramedical}, Huatuo-o1~\citep{chen2024huatuogpt}.
\item \textbf{General Base LLMs:} gemma-3-4b-it (base of MedGemma)~\citep{team2025gemma}, phi-3.5-mini (base of MediPhi)~\citep{abdin2024phi}, Qwen-2.5-7B (base of Meditron3-Qwen2.5)~\citep{yang2025qwen3}, Llama-3.1 (base of Meditron3-Llama3.1, Huatuo-o1, and UltraMedical)~\citep{dubey2024llama}.
\item \textbf{General LLMs:} Qwen3-8B, DeepSeek-R1-Distill-8B~\citep{deepseekai2025deepseekr1incentivizingreasoningcapability}, GPT-4.1-mini, GPT-5~\citep{achiam2023gpt}, gemma-3-27b-it, and Llama-3.3 70B.
\end{itemize}

All models are evaluated using the default inference hyperparameters specified in their original releases. For models that provide both standard and reasoning modes (e.g., Qwen3, Phi-3.5), we evaluate each mode accordingly. Following standard practice in numeracy evaluation, we use zero-shot chain-of-thought prompting (figure~\ref{fig:zero_shot_cot_prompt}) for all experiments, including those run in reasoning mode. 

\subsection{Main Results}
Table~\ref{tab:main-result} reports the performance of LLMs on the four clinical numeracy tasks across a range of clinical context representations. Overall, models perform strongly on retrieval, but substantial challenges remain in clinical numerical reasoning, particularly for comparison and aggregation.

\paragraph{Finding 1: While retrieval succeeds consistently, comparison and aggregation show critical weaknesses.}
Retrieval achieves the highest accuracy across nearly all models: most exceed 85\%, and some reach 100\% (e.g., Gemma, Qwen3, and GPT-4.1-mini). This indicates that LLMs are generally reliable at extracting simple numerical values from clinical contexts. In contrast, arithmetic performance is much more variable, ranging from 18.67\% to 97.50\%. Notably, MedGemma/Gemma performs strongly relative to both larger models (e.g., Llama-3.1 and Huatuo-o1) and same-sized competitors (e.g., MediPhi/Phi-3.5-mini). 
However, with the exception of Qwen3 in reasoning mode and GPT-4.1-mini, most models degrade substantially on comparison and aggregation. Comparison exhibits the largest variance among medical models, while several non-medical reasoning models remain weak on aggregation. Most strikingly, some models achieve near-zero accuracy on comparison (e.g., MedGemma 1.56\% and Meditron3 2.35\%), and Llama-3.1 falls below 30\% on aggregation. These failures suggest that operations requiring global reasoning across multiple values, such as comparison and aggregation, remain a key bottleneck for clinically reliable numeracy.
This dramatic failure mode suggests that comparison and aggregation reasoning, particularly operations requiring global reasoning across multiple values, remains a critical weakness in clinical numerical understanding, even for models that excel at retrieval and basic calculations.

\paragraph{Finding 2: Medical fine-tuning erodes reasoning-based LLMs' numerical reasoning.}
General-purpose LLMs consistently outperform medical-specialized models across all tasks, with the largest gaps on comparison and aggregation. This pattern suggests that medical fine-tuning can reduce numerical reasoning capability. For example, MediPhi attains only 22\% on arithmetic, compared to 75.83\% for its base model Phi-3.5-mini; it also shows large drops on comparison (-23.76\%) and aggregation (-39.3\%). We observe similar trends for other medically adapted models, indicating that standard medical fine-tuning can compromise quantitative reasoning. The extent of degradation also depends on the base model: with a similar fine-tuning strategy, Meditron3 built on Llama-3.1 loses substantially more performance than its Qwen2.5-based counterpart, highlighting the importance of base-model choice. MedGemma further illustrates task-specific trade-offs: arithmetic drops only slightly, but comparison and aggregation decline by 20--60 points relative to its base. In contrast, models fine-tuned with chain-of-thought supervision (UltraMedical and Huatuo-o1) improve over their base models on comparison and aggregation, suggesting that reasoning-trace supervision can preserve or even enhance numeracy during medical adaptation.

\paragraph{Finding 3: Significant performance degradation under variant-template contexts indicates format sensitivity.}
Performance shifts across context representations depend on task complexity and model category. Base models exhibit sizeable robustness gaps between fixed-template and note-style variant contexts (10--22\%), while medical models show smaller but still meaningful declines (0.5--15\%). We also observe a trade-off: Llama-3.1 improves on variant templates relative to fixed templates, whereas CoT-trained medical models (UltraMedical and Huatuo-o1) degrade on variant contexts, suggesting that CoT fine-tuning can boost task performance but reduce format robustness. Even retrieval, despite its high absolute accuracy, can drop by 2--43\% under formatting changes, indicating sensitivity in basic numerical extraction. Comparison shows the greatest robustness variability, with some models suffering catastrophic drops exceeding 22 points on certain variants. The robustness gap between 3 models groups shows that medical LLMs tend to have a lower robustness gap than general LLMs, highlighting the potential impact of format more than the length of the context. Overall, these results suggest that current LLMs lack format-invariant numerical representations and often rely on surface-level cues, which can fail under realistic documentation variation. This raises a critical concern for real-world clinical deployment.

\subsection{Analysis}

\paragraph{Comparison error analysis by question type.}
\begin{figure}[t]
    \centering
    \includegraphics[width=1.\columnwidth]{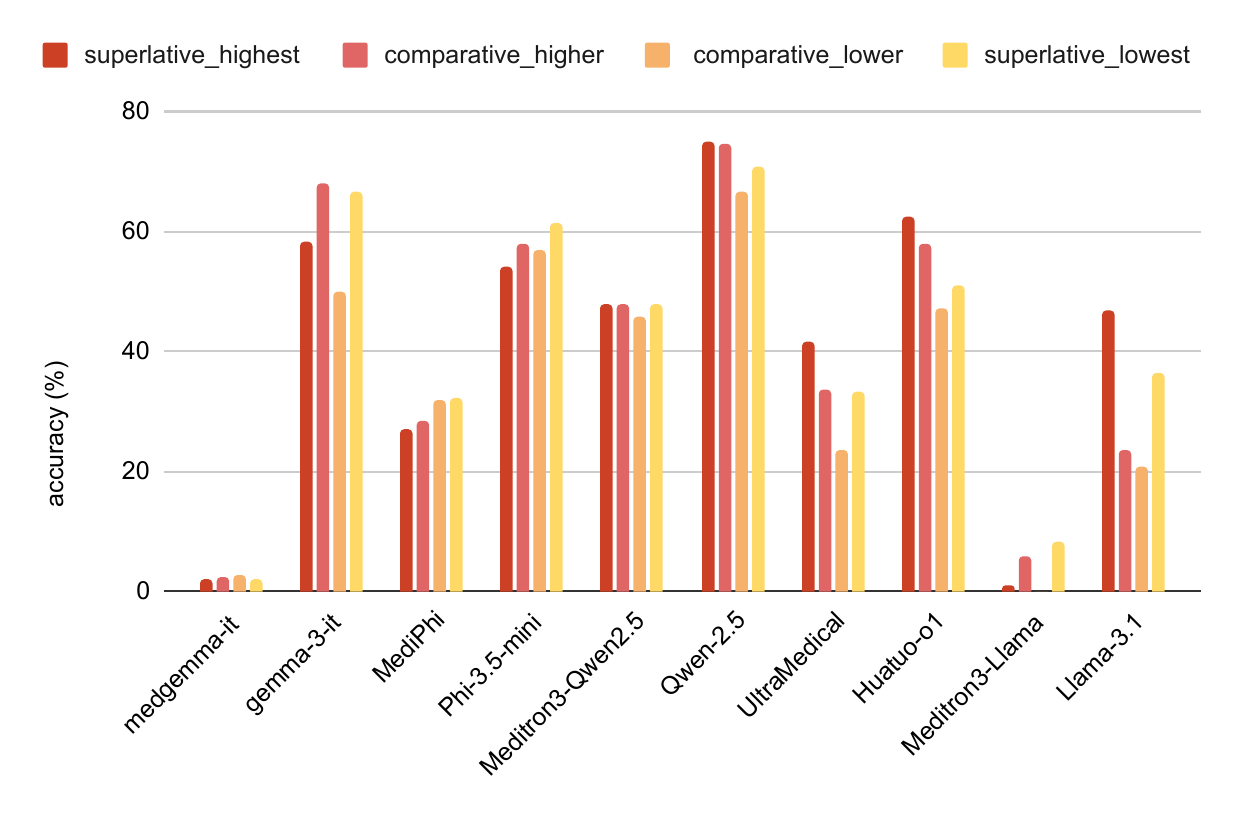}
    \caption{Fine-grained accuracy of comparison question types in fixed-template context.}
    \label{fig:s_fine_grained_comparison}
    \vspace{-1em}
\end{figure}
To better understand failures on comparison, we report accuracy by question type for both comparative and superlative queries: \texttt{highest}, \texttt{higher}, \texttt{lower}, and \texttt{lowest} (see Section~\ref{subsubsec:comparison_tasks} for definitions). Figure~\ref{fig:s_fine_grained_comparison} reports results on the fixed-template context. A consistent pattern emerges: \textit{lower comparative} questions are the most challenging for nearly all models, while the other types show larger model-to-model variation. One plausible reason is that \texttt{lower} comparatives require jointly tracking temporal order and applying a direction-sensitive threshold (e.g., finding the first time a vital drops below a value), which is prone to errors in inequality direction and record selection. In contrast, superlative questions can sometimes be answered by identifying a single extreme value, which may align better with common prompting heuristics. Overall, this breakdown suggests that comparison failures are not uniform, and that threshold-based ``lower-than'' reasoning is a key bottleneck even when models perform well on retrieval and arithmetic.

\paragraph{Aggregation robustness under note-style formatting variations.}
\begin{figure}[t]
    \centering
    \includegraphics[width=.95\columnwidth]{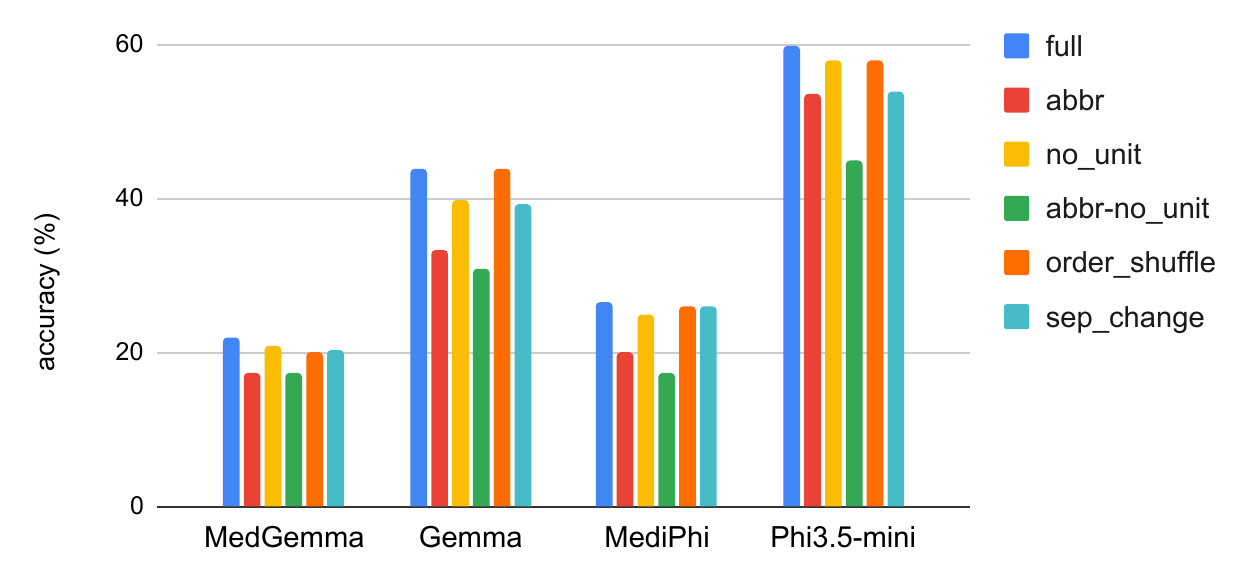}
    \caption{Aggregation accuracy when changing context template configuration. \texttt{full} denotes for the fixed template with full text and unit. \texttt{abbr} denotes for using abbreviations such as bp, hr, rr, o2sat. \texttt{sep\_change} denotes for changing seperator, such as comma, colon.}
    \label{fig:sum_acc_var_change}
    \vspace{-1em}
\end{figure}
To investigate the performance drop under note-style variant contexts, we analyze aggregation accuracy under common documentation variations observed in patient notes. Using templates extracted in Section~\ref{subsubsec:variant_context}, we categorize prevalent surface-form changes, including abbreviations (e.g., BP, HR, RR, O$_2$ sat), separators (e.g., comma, colon, semicolon), and the presence or absence of units (e.g., ``BP 120/70 mmHg'' vs.\ ``BP 120/70''). The distribution of these configurations is summarized in Appendix~\ref{appdx:template_variation_population} (Figure~\ref{fig:template_variant_categorization}). 

Figure~\ref{fig:sum_acc_var_change} reports aggregation accuracy under controlled variants of the fixed-template context, including abbreviation only, unit removal only, and their combination. Results show that abbreviations and missing units are the primary drivers of errors. A plausible explanation is that abbreviations weaken the lexical cues that bind each number to the correct vital-sign attribute, while missing units remove an additional signal for interpreting values and disambiguating fields. When both factors co-occur, errors become more frequent, suggesting that models rely on redundant surface cues and degrade when multiple cues are removed simultaneously. In contrast, changes in attribute order or separator choice have a negligible effect, indicating that models are less sensitive to shallow punctuation changes than to information that directly supports value interpretation and field assignment.

\paragraph{Template substitution fidelity check.}
To ensure that our note-style variants preserve the intended measurements, we validate the extraction-and-substitution pipeline used to instantiate templates. We apply extraction matching to align vital-sign spans and values across variant representations, and \textit{manually} audit 100 randomly sampled outputs for semantic correctness. Specifically, we verify that each substituted vital-sign value corresponds to the correct attribute in the underlying structured record, that units (when present) are consistent with the intended field, and that no attribute is dropped or duplicated after substitution. This audit achieves \textbf{96\% correctness}, indicating that the constructed note-style variants largely reflect the underlying vital-sign records. Therefore, the performance differences observed across context representations are more likely attributable to representational variation rather than template corruption or substitution errors.

\section{Conclusion}
In this paper, we introduce \benchname{}, a benchmark that measures clinical numeracy correctness and robustness to documentation format. Built from longitudinal MIMIC-IV records with programmatic ground truth, \benchname{} evaluates four core numerical operations (retrieval, arithmetic, comparison, and aggregation) under three semantically equivalent context representations. Experiments on 17 LLMs reveal three limitations: comparison and aggregation remain challenging, especially for lower comparative queries; medical fine-tuning can reduce numeracy relative to base models, particularly for comparison; and note-style variation, especially abbreviations and missing units, can substantially degrade performance. Overall, \benchname{} provides a controlled testbed for diagnosing format sensitivity in clinical numerical reasoning and tracking progress toward more reliable numeracy for safer clinical deployment.
\section{Discussion}
Our systematic analysis yields actionable insights for both medical model developers and clinical practitioners. For developers, we identify four critical strategies to enhance model performance: (1) strategic selection of foundational architectures with robust numerical reasoning capabilities, (2) enrichment of training corpora with non-standard clinical formats and abbreviations, (3) augmentation of underrepresented task categories such as comparative reasoning instances, and (4) integration of chain-of-thought supervision to strengthen domain expertise and preserve quantitative reasoning abilities simultaneously. For clinicians, our findings underscore the importance of providing complete contextual information, particularly by including explicit units of measurement and avoiding the concurrent use of abbreviations with unit omissions, which substantially impairs model performance. Additionally, clinicians should exercise heightened caution when deploying these systems for numerical comparison and data aggregation tasks, where performance degradation is most pronounced.
\section*{Ethical Statement}

The data utilized in this study does not contain
any personally identifiable information or offensive content. All data is in English and originates from multiple sources, each distributed under different licenses. The Open Patients dataset \citep{khandekar2024medcalc} is sourced from multiple clinical open access corpora, including the Text Retrieval Conference (TREC) Clinical Decision
Support/Clinical Trials tracks, MedQA-USMLE,
and PMC-Patients. MIMIC-IV \citep{johnson2023mimicorigin} and MIMIC-IV-ED \cite{johnson2023mimic,johnson2024mimic} are publicly available under PhysioNet credentialed access at \href{https://doi.org/10.13026/kpb9-mt58}{https://doi.org/10.13026/kpb9-mt58} and \href{https://doi.org/10.13026/5ntk-km72}{https://doi.org/10.13026/5ntk-km72}, respectively. A condition of releasing a derivative of the MIMIC datasets it to retain the same license. Therefore, the data is submitted and published under PhysioNet credentialed access. 

We used AI assistants (\href{https://manus.im/app}{manus}, \href{https://claude.ai}{Claude}, \href{https://paperreview.ai}{PaperReview}) to refine Figure \ref{fig:full_context_flow}, enhance language, and review the manuscript.
\section*{Limitations}
While our evaluation provides robust insights into LLM capabilities in medical numeracy, there are two primary limitations, which are potential for future investigation. First, our use of question templates enables systematic evaluation of robustness in context understanding and reasoning through controlled variations, ensuring reproducibility and isolated assessment of contextual factors affecting performance. 
However, clinical practice involves stakeholders expressing semantically equivalent queries through unconstrained natural phrasings. Assessing robustness using naturally expressed questions, such as those generated through LLM paraphrasing, LLM generation, or collected from practitioners, would advance understanding of LLM capabilities in capturing the full linguistic variability inherent in clinical settings. Second, we strategically focused on vital signs and related formulas (e.g., shock index, mean arterial pressure), selecting well-established medical concepts present in both structured data (EHRs) and unstructured text (patient notes) that are likely well-represented in LLM training data. This foundation enables confident conclusions about numerical reasoning on core medical concepts. However, the medical knowledge landscape encompasses numerous specialized domains that may be underrepresented in current models, potentially affecting retrieval and reasoning performance. Systematically extending our evaluation framework to these diverse knowledge domains would provide comprehensive understanding of LLM capabilities and limitations across the medical knowledge spectrum, particularly regarding numerical reasoning proficiency.


\bibliography{med-numeracy-eval}

@article{bedi2025medhelm,
  title={{MedHELM: Holistic Evaluation of Large Language Models for Medical Tasks}},
  author={Bedi, Suhana and Cui, Hejie and Fuentes, Miguel and Unell, Alyssa and Wornow, Michael and Banda, Juan M and Kotecha, Nikesh and Keyes, Timothy and Mai, Yifan and Oez, Mert and others},
  journal={{arXiv preprint arXiv:2505.23802}},
  year={2025}
}

@inproceedings{li2025exposing,
  title={{Exposing Numeracy Gaps: A Benchmark to Evaluate Fundamental
Numerical Abilities in Large Language Models}},
  author={Li, Haoyang and Chen, Xuejia and Xu, Zhanchao and Li, Darian and Hu, Nicole and Teng, Fei and Li, Yiming and Qiu, Luyu and Zhang, Chen Jason and Qing, Li and others},
  booktitle={{Findings of the Association for Computational Linguistics: ACL 2025}},
  pages={20004--20026},
  year={2025}
}

@inproceedings{mahendra2024numbers,
  title={{Do numbers matter? Types and prevalence of numbers in clinical texts}},
  author={Mahendra, Rahmad and Spina, Damiano and Cavedon, Lawrence and Verspoor, Karin},
  booktitle={{Proceedings of the 23rd Workshop on Biomedical Natural Language Processing}},
  pages={409--415},
  year={2024}
}

@inproceedings{mahendra2025evaluating,
  title={{Evaluating Numeracy of Language Models as a Natural Language Inference Task}},
  author={Mahendra, Rahmad and Spina, Damiano and Cavedon, Lawrence and Verspoor, Karin},
  booktitle={Findings of the Association for Computational Linguistics: NAACL 2025},
  pages={8336--8361},
  year={2025}
}

@article{khandekar2024medcalc,
  title={{MedCalc-Bench: Evaluating Large Language Models for Medical Calculations}},
  author={Khandekar, Nikhil and Jin, Qiao and Xiong, Guangzhi and Dunn, Soren and Applebaum, Serina and Anwar, Zain and Sarfo-Gyamfi, Maame and Safranek, Conrad and Anwar, Abid and Zhang, Andrew and others},
  journal={{Advances in Neural Information Processing Systems}},
  volume={37},
  pages={84730--84745},
  year={2024}
}

@misc{johnson2024mimic,
  title={{MIMIC-IV (Version 3.1). PhysioNet. RRID: SCR\_007345}},
  author={Johnson, A and Bulgarelli, L and Pollard, T and Gow, B and Moody, B and Horng, S and Celi, LA and Mark, R},
  year={2024}
}

@article{johnson2023mimicorigin,
  title={{MIMIC-IV, a freely accessible electronic health record dataset}},
  author={Johnson, Alistair EW and Bulgarelli, Lucas and Shen, Lu and Gayles, Alvin and Shammout, Ayad and Horng, Steven and Pollard, Tom J and Hao, Sicheng and Moody, Benjamin and Gow, Brian and others},
  journal={Scientific data},
  volume={10},
  number={1},
  pages={1},
  year={2023},
  publisher={{Nature Publishing Group UK London}}
}

@article{johnson2023mimic,
  title={{MIMIC-IV-ED v2.2}},
  author={Johnson, Alistair and Bulgarelli, Lucas and Pollard, Tom and Celi, Leo Anthony and Mark, Roger and Horng, Steven},
  journal={{PhysioNet}},
  year={2023}
}

@article{chen2024huatuogpt,
  title={{HuatuoGPT-o1, Towards Medical Complex Reasoning with LLMs}},
  author={Chen, Junying and Cai, Zhenyang and Ji, Ke and Wang, Xidong and Liu, Wanlong and Wang, Rongsheng and Hou, Jianye and Wang, Benyou},
  journal={arXiv preprint arXiv:2412.18925},
  year={2024}
}

@article{sellergren2025medgemma,
  title={{MedGemma Technical Report}},
  author={Sellergren, Andrew and Kazemzadeh, Sahar and Jaroensri, Tiam and Kiraly, Atilla and Traverse, Madeleine and Kohlberger, Timo and Xu, Shawn and Jamil, Fayaz and Hughes, C{\'\i}an and Lau, Charles and others},
  journal={arXiv preprint arXiv:2507.05201},
  year={2025}
}

@article{yang2025qwen3,
  title={{Qwen3 Technical Report}},
  author={Yang, An and Li, Anfeng and Yang, Baosong and Zhang, Beichen and Hui, Binyuan and Zheng, Bo and Yu, Bowen and Gao, Chang and Huang, Chengen and Lv, Chenxu and others},
  journal={arXiv preprint arXiv:2505.09388},
  year={2025}
}

@article{corbeil2025modular,
    title={{A Modular Approach for Clinical SLMs Driven by Synthetic Data with Pre-Instruction Tuning, Model Merging, and Clinical-Tasks Alignment}},
    author={Corbeil, Jean-Philippe and Dada, Amin and Attendu, Jean-Michel and Abacha, Asma Ben and Sordoni, Alessandro and Caccia, Lucas and Beaulieu, Fran{\c{c}}ois and Lin, Thomas and Kleesiek, Jens and Vozila, Paul},
    journal={arXiv preprint arXiv:2505.10717},
    year={2025}
}

@article{chen2024meditron,
  title={{MEDITRON: Open Medical Foundation Models Adapted for Clinical Practice}},
  author={Chen, Zeming and Romanou, Angelika and Bonnet, Antoine and Hern{\'a}ndez-Cano, Alejandro and Alkhamissi, Badr and Matoba, Kyle and Salvi, Francesco and Pagliardini, Matteo and Fan, Simin and K{\"o}pf, Andreas and others},
  year={2024}
}

@article{team2025gemma,
  title={{Gemma 3 Technical Report}},
  author={Team, Gemma and Kamath, Aishwarya and Ferret, Johan and Pathak, Shreya and Vieillard, Nino and Merhej, Ramona and Perrin, Sarah and Matejovicova, Tatiana and Ram{\'e}, Alexandre and Rivi{\`e}re, Morgane and others},
  journal={arXiv preprint arXiv:2503.19786},
  year={2025}
}

@article{abdin2024phi,
  title={{Phi-4 Technical Report}},
  author={Abdin, Marah and Aneja, Jyoti and Behl, Harkirat and Bubeck, S{\'e}bastien and Eldan, Ronen and Gunasekar, Suriya and Harrison, Michael and Hewett, Russell J and Javaheripi, Mojan and Kauffmann, Piero and others},
  journal={arXiv preprint arXiv:2412.08905},
  year={2024}
}

@article{achiam2023gpt,
  title={{GPT-4 Technical Report}},
  author={Achiam, Josh and Adler, Steven and Agarwal, Sandhini and Ahmad, Lama and Akkaya, Ilge and Aleman, Florencia Leoni and Almeida, Diogo and Altenschmidt, Janko and Altman, Sam and Anadkat, Shyamal and others},
  journal={arXiv preprint arXiv:2303.08774},
  year={2023}
}

@misc{deepseekai2025deepseekr1incentivizingreasoningcapability,
      title={{DeepSeek-R1: Incentivizing Reasoning Capability in LLMs via Reinforcement Learning}}, 
      author={DeepSeek-AI},
      year={2025},
      eprint={2501.12948},
      archivePrefix={arXiv},
      primaryClass={cs.CL},
      url={https://arxiv.org/abs/2501.12948}, 
}

@article{zhang2024ultramedical,
  title={{UltraMedical: Building Specialized Generalists in Biomedicine}},
  author={Zhang, Kaiyan and Zeng, Sihang and Hua, Ermo and Ding, Ning and Chen, Zhang-Ren and Ma, Zhiyuan and Li, Haoxin and Cui, Ganqu and Qi, Biqing and Zhu, Xuekai and others},
  journal={{Advances in Neural Information Processing Systems}},
  volume={37},
  pages={26045--26081},
  year={2024}
}

@article{dubey2024llama,
  title={{The Llama 3 Herd of Models}},
  author={Dubey, Abhimanyu and Jauhri, Abhinav and Pandey, Abhinav and Kadian, Abhishek and Al-Dahle, Ahmad and Letman, Aiesha and Mathur, Akhil and Schelten, Alan and Yang, Amy and Fan, Angela and others},
  journal={arXiv e-prints},
  pages={arXiv--2407},
  year={2024}
}

@inproceedings{nguyen2024carer,
  title={{CARER-ClinicAl Reasoning-Enhanced Representation for Temporal Health Risk Prediction}},
  author={Nguyen, Tuan Dung and Huynh, Thanh Trung and Phan, Minh Hieu and Nguyen, Quoc Viet Hung and Le Nguyen, Phi},
  booktitle={{Proceedings of the 2024 Conference on Empirical Methods in Natural Language Processing}},
  pages={10392--10407},
  year={2024}
}

@article{oliveira2025development,
  title={{Development and evaluation of a clinical note summarization system using large language models}},
  author={Oliveira, Juliana Damasio and Santos, Henrique DP and Ulbrich, Ana Helena DPS and Couto, Julia Colleoni and Arocha, Marcelo and Santos, Joaquim and Costa, Manuela Martins and Faccio, Daniela and Tabalipa, Fabio O and Nogueira, Rodrigo F},
  journal={{Communications Medicine}},
  volume={5},
  number={1},
  pages={376},
  year={2025},
  publisher={Nature Publishing Group UK London}
}

@inproceedings{gourabathina2025medium,
  title={{The Medium is the Message: How Non-Clinical Information Shapes Clinical Decisions in LLMs}},
  author={Gourabathina, Abinitha and Gerych, Walter and Pan, Eileen and Ghassemi, Marzyeh},
  booktitle={{Proceedings of the 2025 ACM Conference on Fairness, Accountability, and Transparency}},
  pages={1805--1828},
  year={2025}
}

@article{cobbe2021training,
  title={{Training Verifiers to Solve Math Word Problems}},
  author={Cobbe, Karl and Kosaraju, Vineet and Bavarian, Mohammad and Chen, Mark and Jun, Heewoo and Kaiser, Lukasz and Plappert, Matthias and Tworek, Jerry and Hilton, Jacob and Nakano, Reiichiro and others},
  journal={arXiv preprint arXiv:2110.14168},
  year={2021}
}

@article{hendrycksmath2021,
  title={{Measuring Mathematical Problem Solving With the MATH Dataset}},
  author={Dan Hendrycks and Collin Burns and Saurav Kadavath and Akul Arora and Steven Basart and Eric Tang and Dawn Song and Jacob Steinhardt},
  journal={{NeurIPS}},
  year={2021}
}

@article{Shao2024DeepSeekMathPT,
  title={{DeepSeekMath: Pushing the Limits of Mathematical Reasoning in Open Language Models}},
  author={Zhihong Shao and Peiyi Wang and Qihao Zhu and Runxin Xu and Jun-Mei Song and Mingchuan Zhang and Y. K. Li and Yu Wu and Daya Guo},
  journal={{ArXiv}},
  year={2024},
  volume={abs/2402.03300},
  url={https://api.semanticscholar.org/CorpusID:267412607}
}

@article{Rothman2008PerspectiveTR,
  title={{Perspective: The Role of Numeracy in Health Care}},
  author={Russell L. Rothman and Victor M. Montori and Andrea L Cherrington and Michael Pignone},
  journal={{Journal of Health Communication}},
  year={2008},
  volume={13},
  pages={583 - 595},
  url={https://api.semanticscholar.org/CorpusID:20118928}
}

@inproceedings{long2025emgllm,
  title={{EMGLLM: Data-to-Text Alignment for Electromyogram Diagnosis Generation with Medical Numerical Data Encoding}},
  author={Long, Zefei and Cao, Zhenbiao and Chen, Wei and Wei, Zhongyu},
  booktitle={{Findings of the Association for Computational Linguistics: ACL 2025}},
  pages={20470--20480},
  year={2025}
}

@article{Ness2024MedFuzzET,
  title={{MedFuzz: Exploring the Robustness of Large Language Models in Medical Question Answering}},
  author={Robert Osazuwa Ness and Katie Matton and Hayden S. Helm and Sheng Zhang and Junaid Bajwa and Carey E. Priebe and Eric Horvitz},
  journal={{ArXiv}},
  year={2024},
  volume={abs/2406.06573},
  url={https://api.semanticscholar.org/CorpusID:270380308}
}

@inproceedings{chen-etal-2025-benchmarking,
    title = "{Benchmarking Large Language Models on Answering and Explaining Challenging Medical Questions}",
    author = "Chen, Hanjie  and
      Fang, Zhouxiang  and
      Singla, Yash  and
      Dredze, Mark",
    editor = "Chiruzzo, Luis  and
      Ritter, Alan  and
      Wang, Lu",
    booktitle = "Proceedings of the 2025 Conference of the Nations of the Americas Chapter of the Association for Computational Linguistics: Human Language Technologies (Volume 1: Long Papers)",
    month = apr,
    year = "2025",
    address = "Albuquerque, New Mexico",
    publisher = "Association for Computational Linguistics",
    url = "https://aclanthology.org/2025.naacl-long.182/",
    doi = "10.18653/v1/2025.naacl-long.182",
    pages = "3563--3599",
    ISBN = "979-8-89176-189-6"
}

@misc{zuo2025medxpertqabenchmarkingexpertlevelmedical,
      title={{MedXpertQA: Benchmarking Expert-Level Medical Reasoning and Understanding}}, 
      author={Yuxin Zuo and Shang Qu and Yifei Li and Zhangren Chen and Xuekai Zhu and Ermo Hua and Kaiyan Zhang and Ning Ding and Bowen Zhou},
      year={2025},
      eprint={2501.18362},
      archivePrefix={arXiv},
      primaryClass={cs.AI},
      url={https://arxiv.org/abs/2501.18362}, 
}

@article{Li2024GSMPlusAC,
  title={{GSM-Plus: A Comprehensive Benchmark for Evaluating the Robustness of LLMs as Mathematical Problem Solvers}},
  author={Qintong Li and Leyang Cui and Xueliang Zhao and Lingpeng Kong and Wei Bi},
  journal={{ArXiv}},
  year={2024},
  volume={abs/2402.19255},
  url={https://api.semanticscholar.org/CorpusID:268063753}
}

@inproceedings{Zhuo2024ProSAAA,
  title={{ProSA: Assessing and Understanding the Prompt Sensitivity of LLMs}},
  author={Jingming Zhuo and Songyang Zhang and Xinyu Fang and Haodong Duan and Dahua Lin and Kai Chen},
  booktitle={{Conference on Empirical Methods in Natural Language Processing}},
  year={2024},
  url={https://api.semanticscholar.org/CorpusID:273375563}
}

@inproceedings{Arora2025ExploringRO,
  title={{Exploring Robustness of LLMs to Paraphrasing Based on Sociodemographic Factors}},
  author={Pulkit Arora and Akbar Karimi and Lucie Flek},
  year={2025},
  url={https://api.semanticscholar.org/CorpusID:275515376}
}

@article{Mehrotra2025UnmaskingBI,
  title={{Unmasking Bias in Financial AI: A Robust Framework for Evaluating and Mitigating Hidden Biases in LLMs}},
  author={Shreshth Mehrotra and Raghavendra P and Balraj Prajesh and Hrishikesh Kambale and Puspita Majumdar},
  journal={{Proceedings of the 6th ACM International Conference on AI in Finance}},
  year={2025},
  url={https://api.semanticscholar.org/CorpusID:283096246}
}

@article{Chen2025CARESCE,
  title={{CARES: Comprehensive Evaluation of Safety and Adversarial Robustness in Medical LLMs}},
  author={Sijia Chen and Xiaomin Li and Mengxue Zhang and Eric Hanchen Jiang and Qin Zeng and Chen-Hsiang Yu},
  journal={ArXiv},
  year={2025},
  volume={abs/2505.11413},
  url={https://api.semanticscholar.org/CorpusID:278715151}
}

@inproceedings{roberts2022overview,
  title={{Overview of the TREC 2022 Clinical Trials Track}},
  author={Roberts, Kirk and Demner-Fushman, Dina and Voorhees, Ellen M and Bedrick, Steven and Hersh, William R},
  booktitle={{TREC}},
  year={2022}
}

@inproceedings{nye2018corpus,
  title={{A Corpus with Multi-Level Annotations of Patients, Interventions and Outcomes to Support Language Processing for Medical Literature}},
  author={Nye, Benjamin and Li, Junyi Jessy and Patel, Roma and Yang, Yinfei and Marshall, Iain and Nenkova, Ani and Wallace, Byron C},
  booktitle={Proceedings of the 56th Annual Meeting of the Association for Computational Linguistics (Volume 1: Long Papers)},
  pages={197--207},
  year={2018}
}

@inproceedings{jin2019pubmedqa,
  title={{PubMedQA: A Dataset for Biomedical Research Question Answering}},
  author={Jin, Qiao and Dhingra, Bhuwan and Liu, Zhengping and Cohen, William and Lu, Xinghua},
  booktitle={{Proceedings of the 2019 Conference on Empirical Methods in Natural Language Processing and the 9th International Joint Conference on Natural Language Processing (EMNLP-IJCNLP)}},
  pages={2567--2577},
  year={2019}
}

@article{jin2020disease,
  title={{What Disease does this Patient Have? A Large-scale Open Domain Question Answering Dataset from Medical Exams}},
  author={Jin, Di and Pan, Eileen and Oufattole, Nassim and Weng, Wei-Hung and Fang, Hanyi and Szolovits, Peter},
  journal={arXiv preprint arXiv:2009.13081},
  year={2020}
}

@article{mccarthy2010mtld,
  title={{MTLD, vocd-D, and HD-D: A validation study of sophisticated approaches to lexical diversity assessment}},
  author={McCarthy, Philip M and Jarvis, Scott},
  journal={{Behavior research methods}},
  volume={42},
  number={2},
  pages={381--392},
  year={2010},
  publisher={Springer}
}

@book{malvern2004lexical,
  title={{Lexical Diversity and Language Development}},
  author={Malvern, David and Richards, Brian and Chipere, Ngoni and Dur{\'a}n, Pilar},
  year={2004},
  publisher={Springer}
}

@article{guiraud1959problemes,
  title={{Probl{\`e}mes et m{\'e}thodes de la statistique linguistique}},
  author={Guiraud, Pierre},
  year={1959}
}

@misc{zhao2021calibrateuseimprovingfewshot,
      title={{Calibrate Before Use: Improving Few-Shot Performance of Language Models}}, 
      author={Tony Z. Zhao and Eric Wallace and Shi Feng and Dan Klein and Sameer Singh},
      year={2021},
      eprint={2102.09690},
      archivePrefix={arXiv},
      primaryClass={cs.CL},
      url={https://arxiv.org/abs/2102.09690}, 
}

\clearpage

\appendix
\label{sec:appendix}
\section{Additional Results}
\label{appendix:results}
\subsection{Arithmetic}
Table~\ref{tab:acc_calc_detail} reports accuracy on arithmetic sub-tasks with increasing computational depth, including 1-step, 2-step, and 3-step calculations. Across models, accuracy generally decreases as the number of required operations increases, indicating that multi-step arithmetic remains substantially harder than single-step computation. Models trained or fine-tuned with reasoning-focused data show smaller performance gaps across depths, suggesting greater robustness to increasing computational complexity. Overall, these results highlight the importance of reasoning-oriented supervision for maintaining stable performance as arithmetic requires more steps.
\begin{table}[h]
\centering
\resizebox{\columnwidth}{!}{
\begin{tabular}{l|ccc}
 \textbf{Model} & \textbf{1-step} & \textbf{2-steps} & \textbf{3-steps} \\
\toprule
MedGemma & 69.50 & 73.50 & 63.50 \\
MediPhi & 21.00 & 14.50 & 14.00 \\
Meditron3-Qwen2.5 & 82.00 & 77.00 & 61.50 \\
Meditron3-Llama & 56.00 & 45.50 & 38.00 \\
UltraMedical & 51.50 & 38.50 & 38.00 \\
Huatuo-o1 & 51.00 & 51.50 & 54.00\\
\hline
Gemma & 79.00 & 72.00 & 68.00 \\
Phi-3.5-mini & 86.50 & 60.00 & 63.00\\
Qwen-2.5 & 90.50 & 78.50 & 78.00 \\
llama-3.1 & 74.00 & 70.00 & 55.00 \\
\hline
DeepSeek-R1-Distill & 65.00 & 64.50 & 27.50 \\
Qwen3 & 95.50 & 84.50 & 68.00\\
GPT-4.1-mini & 96.50 & 92.50 & 88.50 \\
Qwen3-reasoning & 96.00 & 94.00 & 93.00 \\

\end{tabular}
}
\caption{Accuracy of Arithmetic sub-tasks.}
\label{tab:acc_calc_detail}
\end{table}

\subsection{Comparison}
Table~\ref{tab:acc_comparison_detail} reports detailed accuracy for the comparison task across the three context representations.
\begin{table}
\centering
\resizebox{\columnwidth}{!}{%
\begin{tabular}{|l|r|}
\hline
\textbf{LLM}        & \textbf{Model Implementation}    \\
\hline
medgemma-it & google/medgemma-4b-it  \\
gemma-3-it & google/gemma-3-4b-it   \\
MediPhi    & microsoft/MediPhi-Instruct              \\
Phi-3.5-mini &  microsoft/Phi-3.5-mini-instruct         \\
Meditron3-Qwen2.5   & OpenMeditron/Meditron3-Qwen2.5-7B             \\
Meditron3-Llama   & OpenMeditron/Meditron3-8B             \\
UltraMedical   & TsinghuaC3I/Llama-3.1-8B-UltraMedical            \\
Huatuo-o1   &FreedomIntelligence/HuatuoGPT-o1-8B            \\
Qwen-2.5 & Qwen/Qwen2.5-7B-Instruct\\
llama-3.1 & meta-llama/Llama-3.1-8B-Instruct\\
Qwen3 & Qwen/Qwen3-8B\\
DeepSeek-R1-Distill-Llama-8B & unsloth/DeepSeek-R1-Distill-Llama-8B \\
GPT-4.1-mini   & OpenAI API             \\
\hline
\end{tabular}
}
\caption{Details of used LLMs.}
\label{tab:llm-source}
\end{table}
\begin{table*}[h]
\centering
\resizebox{\textwidth}{!}{
\begin{tabular}{l|cccc|cccc|cccc}
\multirow{2}{*}{\textbf{Model}} & \multicolumn{4}{c}{\textbf{Struct.}} & \multicolumn{4}{c}{\textbf{Fixed}} & \multicolumn{4}{c}{\textbf{Variant}} \\
 & \textbf{sup\_highest} & \textbf{com\_higher} & \textbf{com\_lower} & \textbf{ sup\_lowest} & \textbf{sup\_highest} & \textbf{com\_higher} & \textbf{com\_lower} & \textbf{ sup\_lowest} & \textbf{sup\_highest} & \textbf{com\_higher} & \textbf{com\_lower} & \textbf{ sup\_lowest} \\
\toprule
medgemma-it & 3.12 & 1.68 & 1.39 & 0.00 & 2.08 & 2.52 & 2.78 & 2.08 & 2.08 & 0.00 & 2.78 & 2.52\\
gemma-3-it & 61.46 & 69.75 & 55.56 & 63.54 & 58.33 & 68.07 & 50.00 & 66.67 & 46.88 & 46.88 & 30.56 & 36.13 \\
MediPhi & 29.17 & 30.25 & 33.33 & 26.04 & 27.08 & 28.57 & 31.94 & 32.29 & 28.12 & 23.96 & 18.06 & 14.29 \\
Phi-3.5-mini & 53.12 & 54.62 & 52.78 & 52.08 & 54.17 & 57.98 & 56.94 & 61.46 & 44.38 & 47.50 & 48.89 & 46.97 \\
Meditron3-Qwen2.50 & 44.79 & 32.77 & 27.78 & 38.54 & 47.92 & 47.90 & 45.83 & 47.92 & 36.46 & 37.50 & 26.39 & 34.45 \\
Qwen-2.50 & 77.08 & 70.59 & 58.33 & 60.42 & 75.00 & 74.79 & 66.67 & 70.83 & 56.25 & 57.29 & 54.17 & 45.38 \\
UltraMedical & 27.71 & 25.97 & 19.72 & 33.96 & 41.67 & 33.61 & 23.61 & 33.33 & 19.79 & 21.88 & 16.67 & 15.97 \\
Huatuo-o1 & 52.08 & 52.94 & 33.33 & 55.21 & 62.50 & 57.98 & 47.22 & 51.04 & 32.29 & 23.96 & 25.00 & 29.41 \\
Meditron3-Llama & 4.17 & 0.84 & 1.39 & 3.12 & 1.04 & 5.88 & 0.00 & 8.33 & 5.21 & 4.17 & 2.78 & 0.84 \\
Llama-3.1 & 25.62 & 15.04 & 15.56 & 17.29 & 46.88 & 23.53 & 20.83 & 36.46 & 30.62 & 34.79 & 19.17 & 17.73 \\
DeepSeek-R1-Distill-Llama & 61.46 & 73.95 & 59.72 & 57.29 & 60.42 & 62.18 & 52.78 & 66.67 & 37.71 & 43.96 & 39.44 & 48.57 \\
Qwen3-NonReasoning & 76.04 & 66.39 & 59.72 & 81.25 & 80.21 & 75.63 & 68.06 & 73.96 & 50.00 & 50.00 & 45.83 & 51.26 \\
GPT-4.1-mini & 89.58 & 99.16 & 100.00 & 92.71 & 90.62 & 99.16 & 100.00 & 90.62 & 84.38 & 84.38 & 84.72 & 79.83 \\
Qwen3-Reasoning & 89.58 & 99.16 & 97.22 & 92.71 & 90.62 & 97.48 & 100.00 & 93.75 & 72.92 & 81.25 & 79.17 & 76.47 \\

\end{tabular}
}
\caption{Accuracy of comparison sub-tasks. \texttt{sup\_highest} denotes superlative comparisons identifying the highest value, \texttt{sup\_lowest} denotes superlative comparisons identifying the lowest value, \texttt{com\_higher} denotes comparative assessments of whether one value is higher than another, and \texttt{com\_lower} denotes comparative assessments of whether one value is lower than another.}
\label{tab:acc_comparison_detail}
\end{table*}

\begin{table*}[h]
\centering
\resizebox{0.95\textwidth}{!}{
\begin{tabular}{lccccccc}
\toprule
& \textbf{Medical} & \textbf{Qual. Reasoning} & \textbf{Comput.} & \textbf{Non-MCQ} & \textbf{Num.} & \textbf{Granul.} & \textbf{Robust.} \\
\midrule
MedQA & \textcolor{green}{\cmark} & \textcolor{green}{\cmark} & \textcolor{red}{\xmark} & \textcolor{red}{\xmark} & \textcolor{red}{\xmark} & \textcolor{red}{\xmark} & \textcolor{red}{\xmark} \\
MedXpertQA & \textcolor{green}{\cmark} & \textcolor{green}{\cmark} & \textcolor{red}{\xmark} & \textcolor{red}{\xmark} & \textcolor{red}{\xmark} & \textcolor{red}{\xmark} & \textcolor{red}{\xmark} \\
MedBullets & \textcolor{green}{\cmark} & \textcolor{green}{\cmark} & \textcolor{red}{\xmark} & \textcolor{red}{\xmark} & \textcolor{red}{\xmark} & \textcolor{red}{\xmark} & \textcolor{red}{\xmark} \\
GSM8K & \textcolor{red}{\xmark} & \textcolor{green}{\cmark} & \textcolor{green}{\cmark} & \textcolor{green}{\cmark} & \textcolor{red}{\xmark} & \textcolor{red}{\xmark} & \textcolor{red}{\xmark} \\
MATH & \textcolor{red}{\xmark} & \textcolor{green}{\cmark} & \textcolor{green}{\cmark} & \textcolor{green}{\cmark} & \textcolor{red}{\xmark} & \textcolor{red}{\xmark} & \textcolor{red}{\xmark} \\
NumericBench & \textcolor{red}{\xmark} & \textcolor{green}{\cmark} & \textcolor{green}{\cmark} & \textcolor{red}{\xmark} & \textcolor{green}{\cmark} & \textcolor{green}{\cmark} & \textcolor{red}{\xmark} \\
MedCalc & \textcolor{green}{\cmark} & \textcolor{green}{\cmark} & \textcolor{green}{\cmark} & \textcolor{green}{\cmark} & \textcolor{red}{\xmark} & \textcolor{red}{\xmark} & \textcolor{red}{\xmark} \\
\textbf{Ours} & \textcolor{green}{\cmark} & \textcolor{green}{\cmark} & \textcolor{green}{\cmark} & \textcolor{green}{\cmark} & \textcolor{green}{\cmark} & \textcolor{green}{\cmark} & \textcolor{green}{\cmark} \\
\bottomrule
\end{tabular}
}
\caption{Comparison of clinical and numerical reasoning benchmarks for LLM evaluation. Medical: tasks for medical evaluation; Qualitative (Qual) Reasoning: dataset tests qualitative reasoning; Comput.: dataset requires computation (i.e., quantitative reasoning);
Non-MCQ: questions which have a single answer and without the use of multiple choices; Numeracy (Num.): dataset for numeracy evaluation; Granularity (Granul.): dataset test fine-grained levels in particular problem; Robustness (Robust.): the problem space is modeled using several data views/structures.}
\label{tab:benchmark_comparison}
\end{table*}
\begin{table}[h!]
\centering
\small
\resizebox{0.95\columnwidth}{!}{
\begin{tabular}{>{\raggedright\arraybackslash}p{2cm}*{7}{c}}
\toprule
\textbf{Task Category} & 
\rotatebox{90}{\textbf{Measurement}} & 
\rotatebox{90}{\textbf{Temporal}} & 
\rotatebox{90}{\textbf{Ratio/Proportion}} & 
\rotatebox{90}{\textbf{Range}} & 
\rotatebox{90}{\textbf{Frequency}} & 
\rotatebox{90}{\textbf{Ordinal}} & 
\rotatebox{90}{\textbf{Formula/Math}} \\
\midrule



\multicolumn{8}{l}{\textit{\textbf{Retrieval}}} \\
\quad Retrieval &  \checkmark & \checkmark & & & & & \\

\midrule
\multicolumn{8}{l}{\textit{\textbf{Arithmetic}}} \\
\quad 1-step & \checkmark & \checkmark & & & & & \checkmark \\
\quad 2-steps & \checkmark & \checkmark & \checkmark & & & \checkmark & \checkmark \\
\quad 3-steps & \checkmark & \checkmark & \checkmark & \checkmark & & & \checkmark \\

\midrule
\multicolumn{8}{l}{\textit{\textbf{Comparison}}} \\
\quad Comparative & \checkmark & \checkmark & & \checkmark & & & \\
\quad Superlative & \checkmark & \checkmark & & \checkmark & & \checkmark & \\

\midrule
\multicolumn{8}{l}{\textit{\textbf{Aggregation}}} \\
\quad Aggregation & \checkmark & \checkmark & & \checkmark & \checkmark & \checkmark & \checkmark \\


\bottomrule
\end{tabular}
}
\caption{Alignment of benchmark tasks with clinical numerical information types of \citet{mahendra2024numbers}.}
\label{tab:task-type-alignment}
\end{table}

\subsection{Template Variations}
\begin{table*}[h]
\centering
\resizebox{0.95\textwidth}{!}{
\begin{tabular}{l|cccccc}
 \textbf{Model} & \textbf{full} & \textbf{abbr} & \textbf{no\_unit} & \textbf{abbr-no\_unit} & \textbf{order\_shuffle} & \textbf{sep\_change} \\
\toprule
MedGemma & 21.87 & 17.41 & 20.90 & 17.41 & 19.90 & 20.40 \\
Gemma & 43.79 & 33.29 & 39.76 & 30.81 & 43.74 & 39.26 \\
MediPhi & 26.37 & 19.90 & 24.88 & 17.41 & 25.87 & 25.87 \\
Phi3.5-mini & 59.7 & 53.53 & 57.71 & 44.78 & 57.71 & 53.73 \\
Meditron3-Qwen2.5 & 51.74 & 48.26 & 50.74 & 48.25 & 48.26 & 48.76 \\
Qwen-2.5 & 58.21 & 50.25 & 55.72 & 50.72 & 55.72 & 55.22 \\
Meditron3-Llama & 16.92 & 17.41 & 17.91 & 14.43 & 16.42 & 16.42 \\
UltraMedical & 30.85 & 33.83 & 32.84 & 36.82 & 31.84 & 30.35 \\
Huatuo-o1 & 35.32 & 30.85 & 32.84 & 32.84 & 36.82 & 34.33 \\
llama-3.1 & 29.85 & 26.87 & 25.37 & 27.86 & 29.85 & 28.36 \\
DeepSeek-R1-Distill & 34.33 & 32.81 & 33.82 & 27.36 & 34.82 & 34.31 \\
Qwen3 & 76.62 & 75.60 & 74.61 & 74.13 & 75.62 & 74.13\\
GPT-4.1-mini & 82.09 & 80.82 & 81.76 & 79.15 & 81.91 & 81.14 \\
Qwen3-reasoning & 82.59 & 81.08 & 82.08 & 80.60 & 82.08 & 82.09 \\
\end{tabular}
}
\caption{Accuracy of the aggregation task when changing a configuration in the context template. \texttt{full} denotes for the fixed template with full text and unit. \texttt{abbr} denotes for using
abbreviations such as bp, hr, rr, o2sat. \texttt{sep\_change} denotes for changing seperator, such as comma, colon, semicolon.}
\label{tab:acc_summation_change_detail}
\end{table*}

\label{appdx:template_variation_population}
Table~\ref{tab:acc_summation_change_detail} reports accuracy under template variations that reflect common formatting patterns in clinical documentation. While the standardized template uses full attribute names, explicit units, and space separators, real clinical notes exhibit substantial variation. For example, attributes are often abbreviated (e.g., \texttt{heart\_rate} as HR/hr and \texttt{oxygen\_saturation} as O2sat/SpO2), units may be omitted, and separators such as commas or colons are frequently used. These variations capture realistic note formats and enable systematic assessment of model robustness to documentation inconsistencies.

\begin{table*}[ht]
\centering
\small
\begin{tabular}{p{0.08\linewidth} p{0.12\linewidth} p{0.04\linewidth} p{0.5\linewidth} p{0.18\linewidth}}
\toprule
Task & Subtask & ID & Question template & Reasoning focus \\
\midrule
retrieval & retrieval & T1-6 & \texttt{What was the \{column\} of the patient at \{charttime\}?} & Single-record lookup of a vital at a specific timestamp. \\
\midrule

calculation & 1step\_add & T7 & \texttt{Calculate the sum of systolic and diastolic blood pressure at \{charttime\}.} & A step in Mean Arterial Pressure (MAP) \\
calculation & 1step\_sub & T8 & \texttt{Calculate the pulse pressure at \{charttime\}.} & Difference (SBP -- DBP) \\
calculation & 1step\_mul & T9 & \texttt{What is the value when multiplying the temperature at \{charttime\} by 1.8 round to 1 decimal places.} & Scalar multiplication (°C × 1.8) \\
calculation & 1step\_div & T10 & \texttt{Calculate the Shock Index at \{charttime\}, round to 1 decimal places.} & Heart-rate / SBP ratio. \\
calculation & 2step\_add\_div & T11 & \texttt{Calculate the average of respiratory rate at the \{case\} 2 records, round to 1 decimal places.} & Mean of first or last two respiratory-rate entries. \\
calculation & 2step\_sub\_div & T12 & \texttt{Calculate a third of the pulse pressure at \{charttime\}, round to 1 decimal places.} & Pulse-pressure followed by division by three. \\
calculation & 2step\_add\_mul & T13 & \texttt{Convert temperature at \{charttime\} from Celsius to Fahrenheit, round to 1 decimal places.} & Full °C→°F conversion (×1.8 then +32). \\
calculation & 2step\_sub\_mul & T14 & \texttt{Calculate the percentage that heart rate at \{charttime\} exceeds baseline (70 bpm), round to 1 decimal places.} & Percent deviation from a fixed baseline. \\
calculation & 3step\_map & T15 & \texttt{Calculate the Mean Arterial Pressure at \{charttime\}, round to 1 decimal places.} & MAP formula combining SBP and DBP. \\
calculation & 3step\_change & T16 & \texttt{Calculate the percentage change in heart rate from \{charttime\_1\} to \{charttime\_2\}, round to 2 decimal places.} & Relative change across two timestamps. \\
\midrule

comparison & comparative & T17-21 & \texttt{What is the record datetime when \{vital\_parameter\} first exceed \{threshold\}?} & Detect first crossing of clinician-defined thresholds. \\
comparison & comparative & T22-27 & \texttt{What is the record datetime when \{vital\_parameter\} first drop below \{threshold\}?} & Detect first crossing of clinician-defined thresholds. \\
comparison & superlative & T28-33 & \texttt{What is the record datetime when \{vital\_parameter\} is highest?} & Identify extremal measurement max timestamps. \\
comparison & superlative & T34-39 & \texttt{What is the record datetime when \{vital\_parameter\} is lowest?} & Identify extremal measurement min timestamps. \\
\midrule

aggregation & aggregation & T40-42 & \texttt{How many time the patient has \{issue\}? Return the number of records.} & Count events such as tachycardia, MAP > 100\,mmHg, shock index > 0.7. \\
\bottomrule
\end{tabular}
\caption{Complete set of question templates used in the MIMIC-IV-ED Med Numeracy preprocessing pipeline.}
\label{tab:question_templates}
\end{table*}



\section{Benchmark Comparisons}
\label{appdx:benchmarks_comparison}
We compare \benchname to existing benchmarks in Table~\ref{tab:benchmark_comparison}. Table~\ref{tab:benchmark_comparison_medcalc_detail} provides a detailed comparison with MedCalc-Bench, which also evaluates numerical reasoning but focuses primarily on arithmetic computation.

\section{Lexical Diversity}
\label{appdx:lexical_diversity}
\begin{table}[h]
\centering
\resizebox{0.95\columnwidth}{!}{
\begin{tabular}{l|c|c|c}
\textbf{Task} & \textbf{Structured} & \textbf{Fixed-Tem.} & \textbf{Var-Tem.} \\
\toprule

Retrieval & 4.9268 & 4.3826 & 12.5604 \\
Calculation & 5.3545 & 5.5130 & 12.5724 \\
Comparison & 5.4663 & 4.6588 & 12.8695 \\
Aggregation & 5.4866 & 4.7530 & 12.8291 \\
\bottomrule
\end{tabular}
}
\caption{The lexical diversity of questions in three context representations using Root TTR \citep{guiraud1959problemes}}
\label{tab:lexical_diversity}
\end{table}
\begin{table}[h]
\centering
\resizebox{0.95\columnwidth}{!}{
\begin{tabular}{l|c|c}
\textbf{Metric} & \textbf{MedCalc} & \textbf{Var-Tem.} \\
\toprule
Root TTR \citep{guiraud1959problemes} & 10.7108 & 12.7079\\
HDD \citep{malvern2004lexical} & 0.7335 & 0.8487\\
MTLD \citep{mccarthy2010mtld} & 96.0998 & 84.6647\\

\bottomrule
\end{tabular}
}
\caption{The lexical diversity of questions between MedCalc and our dataset}
\label{tab:lexical_diversity_medcalc}
\end{table}
To quantify the linguistic variation across our context representations, we measure lexical diversity for each format and compare our most realistic setting against MedCalc-Bench \citep{khandekar2024medcalc}. Following standard practice in corpus linguistics, we use the root type--token ratio (Root TTR) \citep{guiraud1959problemes}:
\begin{equation}
\text{Root\_TTR} = \frac{\text{Number of unique tokens}}{\sqrt{\text{Total number of tokens}}}.
\end{equation}
Higher Root TTR indicates greater vocabulary richness and surface-form variability.

\paragraph{Internal comparison across context formats.}
Tables~\ref{tab:lexical_diversity} and~\ref{tab:lexical_diversity_medcalc} summarize internal and external comparisons, respectively. Internally, the note-style variant representation (Var-Tem.) exhibits substantially higher lexical diversity, with Root TTR ranging from 12.56 to 12.87, compared to structured (4.93--5.49) and fixed-template (4.38--5.51) representations. This makes Var-Tem.\ about two to three times more lexically diverse than the two controlled baselines, reflecting the note-derived templates used in Var-Tem., including abbreviations, unit omissions, and formatting differences that mirror real documentation practices.

\paragraph{External comparison with MedCalc-Bench.}
Relative to MedCalc, our Var-Tem.\ contexts have Root TTR that is about two points higher (12.71 vs.\ 10.71) and HDD that is about 0.12 higher (0.85 vs.\ 0.73), indicating greater local vocabulary variation. By contrast, MedCalc-Bench attains higher MTLD (96.10 vs.\ 84.66), suggesting lexical diversity is maintained over longer passages. This divergence is expected because Root TTR and HDD emphasize local surface variation introduced at the record level, whereas MTLD captures sustained diversity across extended text and is influenced by MedCalc-Bench's longer narrative note style.

Overall, these metrics confirm that \benchname{} spans a controlled spectrum of linguistic complexity, from structured and fixed-template baselines to a note-style variant that better reflects real clinical documentation variability.

\begin{table}[h]
\centering
\resizebox{0.95\columnwidth}{!}{
\begin{tabular}{l|c}
\textbf{Benchmarks \& Datasets} & \textbf{\%Num} \\
\toprule
MedXpertQA \citep{zuo2025medxpertqabenchmarkingexpertlevelmedical} & 73.51 \\
Medbullets \citep{chen-etal-2025-benchmarking} & 99.35 \\
MedQA \citep{jin2020disease} & 85.55 \\
MedCalc-Bench \citep{khandekar2024medcalc} & 96.73 \\
PubMedQA* \citep{jin2019pubmedqa} & 96.50 \\
\hline
EBM-NLP* \cite{nye2018corpus} & 90.00 \\
TREC-CDS* \citep{roberts2022overview} & 100.00 \\
\bottomrule

\end{tabular}
}
\caption{The ratio of samples containing numerical information in question contexts, articles, or patient notes. * denotes for the value from \citet{mahendra2024numbers}.}
\label{tab:ratio_num_in_challenging_benchmark}
\end{table}
\begin{table}[h]
\centering
\resizebox{0.95\columnwidth}{!}{
\begin{tabular}{l|l|c|c}
\multicolumn{2}{c|}{} & \textbf{MedXpertQA} & \textbf{ MedBullets} \\
\hline
\multicolumn{2}{l|}{Full data} & 13.06 & 48.70 \\
\cline{2-4}
\multicolumn{2}{l|}{Number-only} & 12.72 & 48.69 \\
\hline
\multicolumn{2}{l|}{\%NumSamples} & 73.51\% & 99.35\% \\
\end{tabular}
}
\caption{The accuracy of 2 challenging clinical benchmarks. \%NumSample denotes the ratio of samples containing at least 2 numbers as numerical information in question. Number-only denotes the accuracy of these samples.}
\label{tab:acc_challenging_benchmark}
\end{table}

\begin{figure*}[t]
    \centering
    \includegraphics[width=\textwidth]{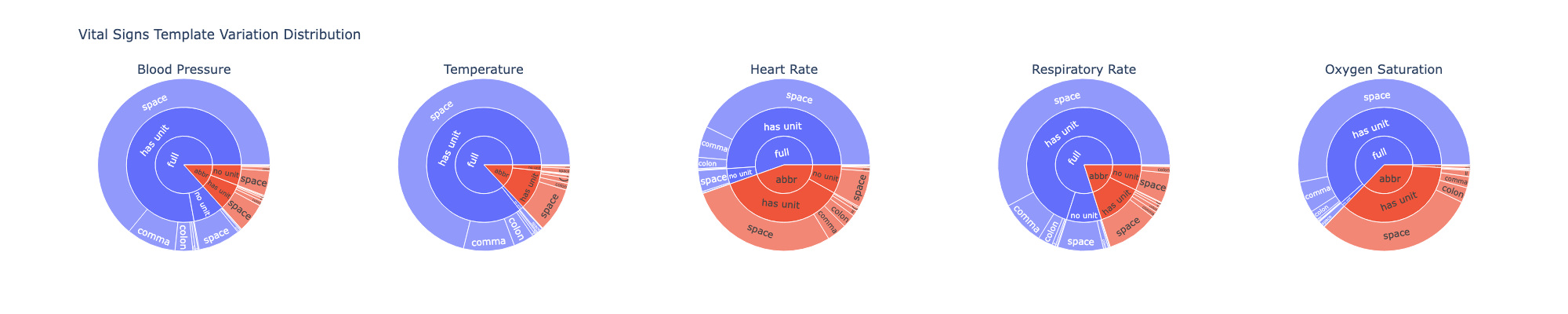}
    \caption{The distribution of template variations. \texttt{full} denotes full texting of vital sign name, such as \textit{blood pressure}, opposite to \texttt{abbr} using abbreviations such as \textit{bp}. The lightest circle denotes the separator between the vital sign name and its value used in the template}\label{fig:template_variant_categorization}
\end{figure*}

\section{The Prevalence of Numerical Information in Clinical Settings}
\citet{mahendra2024numbers} documents the prevalence of numerical information in clinical data. To assess the impact of numerical content on model performance, we conduct extensive evaluation on challenging benchmarks including MedXpertQA \citep{zuo2025medxpertqabenchmarkingexpertlevelmedical} and MedBullets \citep{jin2020disease}. Table~\ref{tab:ratio_num_in_challenging_benchmark} shows the prevalence of numerical data across these benchmarks, while Table~\ref{tab:acc_challenging_benchmark} presents corresponding accuracy results. Our analysis reveals that numerical information appears in 73.51--100\% of questions across challenging medical benchmarks and real-world data, yet model performance drops around 0.3\% on number-containing questions compared to low overall accuracy, revealing a critical weakness in current LLMs and the need of attention on clinical numeracy.

\section{Prompts}

\begin{figure}
    \begin{minipage}{0.9\columnwidth}
        \centering
        \begin{tcolorbox}[title=The Zero-shot CoT Prompt, fonttitle=\bfseries\small]
            You are a helpful clinical assistant in \texttt{{<task>}}. Please think step-by-step to solve the question. Your final output should end with a JSON dict formatted as \{"answer": <short\_and\_direct\_answer\_of\_the\_\\question>\}.\\

            Here is the task:\\
            {<question>}
            \\
            \\
            Let's think step-by-step to solve the question.
        \end{tcolorbox}
      \vspace{1mm}
    \end{minipage}
    \caption{The zero-shot CoT prompt used for evaluation. \texttt{{<task>}} denotes the evaluating task, which is retrieval/arithmetic/comparison/aggregation.}
    \label{fig:zero_shot_cot_prompt}
\end{figure}
\begin{figure*}[h]
    \begin{minipage}{0.95\textwidth}
        \centering
        \begin{tcolorbox}[title=Extraction Prompt, fonttitle=\bfseries\small]
You are a medical documentation specialist. Your task is to extract raw vital sign text in given patient note which contains temperature, heart rate, respiratory rate, oxygen saturation, blood pressure. Text extracted must be a exact match from the patient note.
Output format follow the JSON schema:
\begin{lstlisting}[
  basicstyle=\small\ttfamily,
  breaklines=true,
  columns=fullflexible,
  keepspaces=true
]
{
  "$defs": {
    "ValueObj": {
      "properties": {
        "text": {"type": ["string", "null"], 
                 "description": "Raw text of the value"},
        "number": {"type": ["string", "null"], 
                   "description": "Number of the value"},
        "unit": {"type": ["string", "null"], 
                 "description": "Unit of the value"}
      },
      "required": ["text", "number", "unit"]
    }
  },
  "properties": {
    "temperature": {"$ref": "#/$defs/ValueObj"},
    "heart_rate": {"$ref": "#/$defs/ValueObj"},
    "respiratory_rate": {"$ref": "#/$defs/ValueObj"},
    "oxygen_saturation": {"$ref": "#/$defs/ValueObj"},
    "blood_pressure": {"$ref": "#/$defs/ValueObj"}
  },
  "required": ["temperature", "heart_rate", "respiratory_rate",
               "oxygen_saturation", "blood_pressure"]
}
\end{lstlisting}
\vspace{1em}
Patient Note:\\
\texttt{<patient\_note>}
        \end{tcolorbox}
      \vspace{1mm}
    \end{minipage}
    \caption{Extraction Prompt used to extract a match from patient notes for mapping variant templates of the variant context representation.}
    \label{fig:extraction_prompt}
\end{figure*}

\end{document}